\documentclass[11pt]{article}

\usepackage{microtype}
\usepackage[margin=1in]{geometry}
\usepackage{geometry}
\usepackage{graphicx}
\usepackage{subcaption}
\usepackage{booktabs} %
\usepackage{mathpazo}
\usepackage{pifont}
\usepackage{makecell}
\usepackage{tabularx}
\usepackage{pbox}
\usepackage{soul}

\usepackage[dvipsnames]{xcolor}
\usepackage{pgfplots}
\usepackage{pgfplotstable}
\pgfplotsset{compat=1.3}
\usepackage{tikz}
\usetikzlibrary{pgfplots.groupplots}
\usepackage{ragged2e}

\usepackage[toc,page,header]{appendix}
\usepackage{minitoc}

\usepackage[colorlinks=true,linkcolor=blue,urlcolor=blue,citecolor=blue]{hyperref}
\usepackage{xspace}
\usepackage{bm}
\usepackage{comment}
\usepackage{caption}
\usepackage{makecell}

\usepackage{natbib}
\newcommand{\norm}[1]{\left\lVert#1\right\rVert}
\newcommand{\bx}{\textbf{x}}
\newcommand{\bz}{\textbf{z}}
\newcommand{\beps}{\bm{\varepsilon}}
\newcommand{\trak}{\textsc{trak}\xspace}

\newcommand{\mscoco}{\textsc{MS COCO}\xspace}

\newcommand{\thetastar}{\theta^\star}

\renewcommand{\hat}[1]{\widehat{#1}}

\newlength\myindent
\usepackage[normalem]{ulem}

\usepackage{amsmath}
\usepackage{amssymb}
\usepackage{mathtools}
\usepackage{amsthm}
\usepackage{thmtools}
\usepackage{algorithm}
\usepackage{algpseudocode}
\usepackage[T1]{fontenc}
\usepackage{color}

\usepackage[capitalize,noabbrev]{cleveref}

\definecolor{mydarkblue}{rgb}{0,0.08,0.85}
\definecolor{mylightblue}{rgb}{0.06,0.56,1.0}
\definecolor{mylightorange}{rgb}{1.0,0.62,0.12}
\definecolor{mylightred}{rgb}{0.99,0.00,0.04}
\hypersetup{colorlinks=true,
linkcolor=mydarkblue,
citecolor=mydarkblue,
filecolor=mydarkblue,
urlcolor=mydarkblue,
bookmarksopen=true,
linktoc=page}

\theoremstyle{plain}

\theoremstyle{definition}

\theoremstyle{remark}

\usepackage{marginnote}

\usepackage{todonotes}

\title{The Journey, Not the Destination:\\ How Data Guides Diffusion Models}
\author{
\normalsize
Kristian Georgiev\footnote{Equal contribution.},\ \,
Joshua Vendrow\footnotemark[1],\ \,
Hadi Salman,\ \,
Sung Min Park,\ \,
Aleksander M\k{a}dry \\
\normalsize MIT\\
\texttt{\{krisgrg,jvendrow,hady,sp765,madry\}@mit.edu}}

\date{}

\makeatletter
\let\c@figure\c@table
\makeatother

\begin{document}
\setcounter{tocdepth}{2}
\doparttoc %
\renewcommand\ptctitle{}
\faketableofcontents %

\maketitle
\begin{abstract}
  Diffusion models trained on large datasets can synthesize photo-realistic images
of remarkable quality and diversity. However, {\em attributing} these images
back to the training data—that is, identifying specific training examples which
{\em caused} an image to be generated—remains a challenge. In this paper, we
propose a framework that: (i) provides a formal notion of data attribution in
the context of diffusion models, and (ii) allows us to \textit{counterfactually}
validate such attributions.  Then, we provide a method for computing these
attributions efficiently.  Finally, we apply our method to find (and evaluate)
such attributions for denoising diffusion probabilistic models trained on
CIFAR-10 and latent diffusion models trained on MS~COCO. We provide code at
\href{https://github.com/MadryLab/journey-TRAK}{this https URL}.

\end{abstract}

\section{Introduction}
\label{sec:intro}
Diffusion models can generate novel images that are simultaneously
photorealistic and highly controllable via textual prompting
\citep{ramesh2022hierarchical,rombach2022high}.  A key driver of diffusion
models' performance is training them on massive amounts of data
\citep{schuhmann2022laion}. Yet, this dependence on data has given rise to concerns about
how diffusion models use it.

{For example,
\citet{carlini2021extracting,somepalli2022diffusion} show that diffusion models
often memorize training images and ``regurgitate'' them during generation.
However, beyond such cases of direct memorization, we currently lack a method
for {\em attributing} generated images back to the most influential training
examples---that is, identifying examples that {\em caused} a given image to be
generated.
Indeed, such a primitive---a {\em data attribution method}---would have a number of applications.
For example, previous work has shown that attributing model outputs back
to data can be important for debugging model behavior~\citep{shah2022modeldiff},
detecting poisoned or mislabelled data \citep{lin2022measuring}, and curating
higher quality training datasets \citep{khanna2019interpreting}.
Within the context of diffusion models, data attribution can also help detect cases of
data leakage (i.e., privacy violations),
and more broadly,
can be a valuable tool in the context of tracing content
provenance relevant to questions of
copyright~\citep{andersen2023class,images2023getty}.
Finally, synthetic images generated by diffusion models are now increasingly used across the entire machine learning
pipeline, including training \citep{azizi2023synthetic} and model evaluation
\citep{kattakinda2022invariant,wiles2022discovering,vendrow2023dataset}.  Thus,
it is critical to identify (and mitigate) failure modes of these models
that stem from training data, such as bias propagation
\citep{luccioni2023stable,perera2023analyzing} and memorization.
Motivated by all the above needs, we thus ask:

\begin{figure*}[!t]
    \centering
    \includegraphics[width=\linewidth]{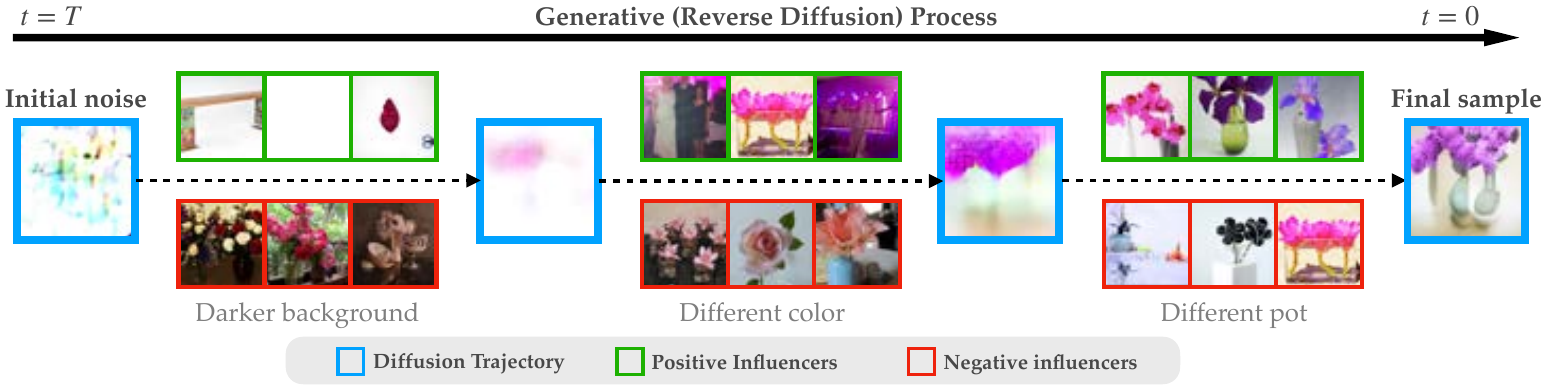}
    \caption{\textbf{Overview of our attribution framework.} For a given
    synthesized image, we apply our attribution method at individual steps along
    the diffusion trajectory. At each step $t$, our method pinpoints the
    training examples with the highest influence (positive in {green}, negative
    in red) on the generative process at that step. In particular, positive
    influencers guide the trajectory towards the final sample, while negative
    influencers guide the trajectory away from it. We observe that negative
    influencers increasingly resemble the final sample (the grey text highlights
    notable differences with the final sample).  For more examples, see Appendix~\ref{app:omitted}.}
    \label{fig:hero_fig}
\end{figure*}

\begin{center}
    {\em How can we reliably attribute images synthesized by diffusion models
    back to the training data?}
\end{center}
\noindent Although data attribution has been extensively studied in the context
of {\em supervised} learning
\citep{koh2017understanding,ghorbani2019towards,jia2019towards,ilyas2022datamodels,hammoudeh2022training,park2023trak},
the generative setting poses new challenges. First, it is unclear {\em what
particular behavior} of these models we hope to attribute. For example, given a
generated image, certain training images might be responsible for the look of
the background, while others might be responsible for the choice of an object
appearing in the foreground.  Second, it is not immediately obvious how to
\textit{verify} the attributions. In supervised settings, a standard approach is
to compare the outputs of the original model on given inputs with those of a new
model trained on a new dataset after removing the attributed examples.  However,
in the generative setting it is less clear how to make such comparisons.

\paragraph{Our contributions.}
In this work, we present a data attribution framework for diffusion models.
This framework reflects, and is motivated by, the fact that diffusion models
iteratively denoise an initial random seed to generate the final image.  In
particular, rather than attributing {\em only} the final generated image, i.e.,
the ``destination,'' we attribute each individual step along the (denoising)
``journey'' taken by diffusion model (see \cref{fig:hero_fig}). This approach
shifts our focus from the specific final image to the {\em distribution} of
possible generated images and, in particular, how this distribution evolves
across the diffusion process. As we demonstrate, this framework also enables us to attribute specific {\em features} of
the final generated image.

To analyze this framework, we introduce two
complementary metrics for evaluating the resulting attributions based on their
counterfactual impact on the distribution of generated images (rather than on specific samples).  Finally, we
provide an efficient method for computing such attributions, building on data
attribution approaches developed for the supervised
setting~\citep{ilyas2022datamodels,park2023trak}. We then apply our method to
denoising diffusion probabilistic models (DDPM)~\citep{ho2020denoising} trained
on CIFAR-10~\citep{krizhevsky2009learning}, and latent diffusion models
(LDM)~\citep{rombach2022high} trained on MS~COCO~\citep{lin2014microsoft}.  In
both of these settings, we obtain attributions that are validated by our
metrics and also visually interpretable.

\section{Preliminaries}
\label{sec:background}
We first provide background on data attribution. Then, we give a brief overview of diffusion models, highlighting the components that we will need to
formalize attribution for these models.

\subsection{Data attribution}
\label{sub:background_attribution}
Broadly, the goal of training data attribution
\citep{koh2017understanding,ilyas2022datamodels,hammoudeh2022training,park2023trak}
is to trace model outputs back to the training data.  Intuitively, we want to
estimate how the presence of each example in the training set impacts a
given model output of interest (e.g., the loss of a classifier) on a specific
input.

To formalize this, consider a learning algorithm $\mathcal{A}$ (e.g., a
training recipe for a model), together with an input space $\mathcal{Z}$ and a
training dataset $S=(z_1, \ldots, z_n)\in\mathcal{Z}^n$ of $n$ datapoints from that input space. Given a datapoint $z\in
\mathcal{Z}$, we represent the model output via a \textit{model output function}
$f(z, \theta(S)):\mathcal{Z}\times\mathbb{R}^d\to \mathbb{R}$, where
$\theta(S)\in\mathbb{R}^d$ denotes the model parameters resulting from running
algorithm $\mathcal{A}$ on the dataset $\mathcal{S}$.  For example, $f(z,
\theta(S))$ is the loss on a test sample $z$ of a classifier trained on $S$. (
Our notation here reflects the fact that the parameters are a function of
the training dataset $S$.)
We now define a {\em data attribution method} as a function $\tau\colon
\mathcal{Z}\times\mathcal{Z}^n\rightarrow \mathbb{R}^n$ that assigns a score
$\tau(z, S)_i\in\mathbb{R}$ to each training example $z_i \in S$.\footnote{Following the literature, we say that an example $z_i$ has a {\em positive (respectively, negative) influence} if $\tau(z,S)_i > 0$ (respectively, $\tau(z,S)_i < 0$).} Intuitively,
we want $\tau(z, S)_i$ to capture the change in the model output function $f(z,
\theta(S))$ induced by adding $z_i$ to the training set.

More generally, these scores should help us make {\em counterfactual} predictions about the model behavior resulting from training on an arbitrary subset $S'\subseteq S$ of the training datapoints.
We can formalize this goal using the {\em datamodeling} task \cite{ilyas2022datamodels}: given an arbitrary subset $S' \subseteq S$ of the training set, the task is to predict the resulting model output $f(z, \theta(S'))$.
A simple method to use the attribution scores for this task, then, is to consider a {\em linear} predictor: $f(z,\theta(S')) \approx \sum_{i: z_i\in S'}\tau(z, S)_i$.\footnote{Similarly to the prior work \citep{park2023trak}, we only consider linear predictors here. }

This view of the data attribution as a prediction task motivates a natural
metric for evaluating attribution methods: the agreement between the true output
$f(z,\theta(S'))$ and the output predicted by the attribution method $\tau$.
\citet{park2023trak} consider the rank correlation between the true and predicted values of $f(z,\theta(S'))$ over different random samples
$S' \subseteq S$ and name the corresponding metric the {\em linear datamodeling
score}---we will adapt it to our setting in \Cref{sec:prelim}.

\paragraph{Estimating attribution scores (efficiently).}
Given the model output function $f$ evaluated at input $z$, a natural way to
assign an attribution score $\tau(z)_i$ for a training datapoint $z_i$ is to
consider the {\em marginal} effect of including that particular example on the
model output, i.e., have
$
    \tau(z)_i = f(z,\theta(S)) - f(z,\theta(S\setminus \{z_i\})).
$
We can further approximate this difference by decomposing it as:

\begin{align}
    \tau(z)_i = \underbrace{(\theta-\theta_{-i})}_\text{(i) change in model
    parameters} \cdot \overbrace{\nabla_\theta f(z,\theta)}^\text{(ii) change in
    model output}, \label{eq:attribution}
\end{align}
where $\theta_{-i}$ denotes $\theta(S \setminus \{i\})$
\citep{wojnowicz2016influence,koh2017understanding}.
We can compute the second component efficiently, as this only requires taking the gradient of the model output function with respect to the parameters; in contrast, computing the first component is not always straightforward. In simpler settings, such as linear regression, we can compute the first
component explicitly, as there exists a closed-form solution for computing the
parameters $\theta(S')$ as a function of the training set $S'$.  However, in modern, non-convex settings,
estimating this component efficiently (i.e., without re-training the model)
is challenging. Indeed, prior works such as
influence functions \citep{koh2017understanding} and TracIn
\citep{pruthi2020estimating} estimate the change in model parameters using
different heuristics, but these approaches can be inaccurate in such settings.

To address these challenges, \trak \citep{park2023trak} observed that for deep
neural networks, approximating the original model with a model that is {\em
linear} in its parameters, and averaging the estimates over multiple $\theta$'s
(to overcome stochasticity in training) yields highly accurate attribution
scores.  The linearization is motivated by the observation that at small
learning rates, the trajectory of gradient descent on the original neural
network is well approximated by that of a corresponding linear model
\citep{long2021properties,wei2022more,malladi2022kernel}.  In this paper, we
will leverage the \trak framework towards attributing diffusion models.

\subsection{Diffusion models}
\label{sub:background_diffusion}

\paragraph{Training and sampling from diffusion models.} At a high level,
diffusion models (and generative models, more broadly) learn a distribution
$p_{\theta}(\cdot)$ meant to approximate a target distribution $q_{data}(\cdot)$
of interest (e.g., natural images).  To perform such learning, given a
(training) sample $\bx_0\sim q_{\text{data}}(\cdot)$, diffusion models first
apply a stochastic \textit{diffusion process} that gradually corrupts $\bx_0$ by
adding more noise to it at each step. This results in a sequence of intermediate
latents $\{\bx_t\}_{t\in[T]}$ sampled according to $\bx_t\sim
\mathcal{N}\left(\alpha_t\cdot \bx_{t-1}, (1-\alpha_t)\cdot I\right)$ where
$\{\alpha_t\}_t$ are parameters of the diffusion process
\citep{sohldickstein2015deep,song2019generative,ho2020denoising}.  Then, based
on such sequences of intermediate latents, diffusion models learn a
``denoising'' neural network $\bm{\varepsilon_{\theta}}$ that attempts to run
the diffusion process ``in reverse.''

Once such a diffusion model is trained, one can sample from it by providing that model with an initial seed $\bx_T\sim\mathcal{N}\left(0,
1\right)$ (i.e., just a sample of random noise), and then applying the (trained) denoising network iteratively at each step $t$ (from $t=T$
to $t=0$) to sample the corresponding \textit{diffusion trajectory}
$\{\bx_t\}_{t\in[T]}$, ultimately leading to a final sample $\bx_0\sim
p_{\theta}(\cdot)\approx q_{data}(\cdot)$.

\paragraph{Conditioning sampling on partially denoised images.}
Importantly, in this work, it will be also useful to consider the process of
sampling a final image $\bx_0$ when ``resuming'' the diffusion process after
running it up to some step $t$---this is equivalent to continuing that process
at step $t$ from the corresponding intermediate latent $\bx_t$. We denote the
distribution arising from sampling an image $\bx_0$ when conditioning on the
latent $\bx_t$ by $p_{\theta}(\cdot | \bx_t)$.

Also, it turns out that we can approximate the multi-step denoising process of
generating samples from $p_\theta(\cdot | \bx_t)$ in a single step with the formula
    $
    \hat{\bx}_0^{t} :=c_1(\alpha_t)\cdot \left(\bx_t - c_2(\alpha_t\cdot \bm{\varepsilon}_\theta(\bx_t, t))\right),
    $
for some constants $c_1(\cdot), c_2(\cdot)$ that depend on the diffusion
parameters $\{\alpha_t\}_t$ \citep{ho2020denoising}. In fact,
$\hat{\bx}_0^{t}$ is a proxy for the conditional expectation
$\mathbb{E}_{\bx_0 \sim p_{\theta}(\cdot|\bx_t)}[\bx_0]$,
and under certain conditions $\hat{\bx}_0^{t}$ is precisely equivalent to this
expectation \citep{song2023consistency,daras2023consistent}.\footnote{This equivalence is referred to as the {\em consistency}
property.} See
\cref{fig:diffusion_background} for an illustration of $p_\theta(\cdot | \bx_t)$
and $\hat{\bx}_0^{t}$ for different values of $t$.

\paragraph{Types of diffusion models.} Finally, Denoising Diffusion
Probabilistic Models (DDPMs) are a popular instantiation of diffusion
models~\citep{ho2020denoising}. More recently, \citet{rombach2022high} proposed
a new class of diffusion models called latent diffusion models (LDMs), which
perform the above stochastic process in the latent space of a pretrained encoder
network. Moreover, \citet{song2021score,ho2022classifier} show that one can also
{\em condition} diffusion models on some additional information, e.g. a text
prompt. This way, one can control the semantics of the generated images by
specifying such a text prompt. In this work, we will instantiate our data attribution
framework on both unconditional DDPMs and conditional LDMs.

\begin{figure}[t!]
    \centering
    \includegraphics[width=0.98\textwidth]{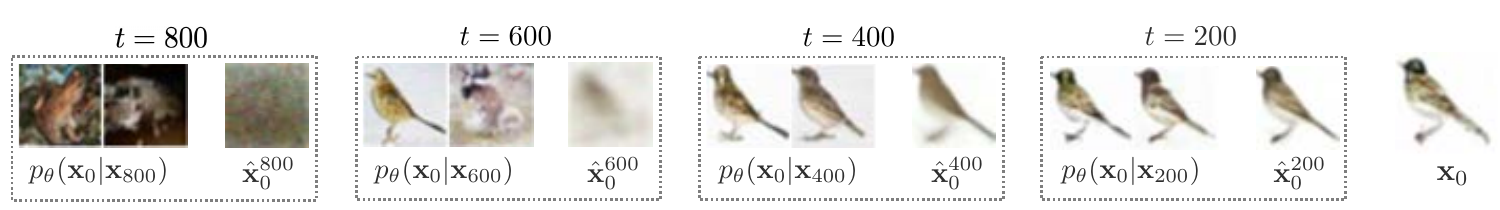}
    \caption{
        {\bf Samples from a diffusion trajectory.} We show samples from $p_{\theta}(\cdot | \bx_t)$, i.e., the distribution of
        final images $\bx_0$ conditioned on initializing from the latent $\bx_t$
        at step $t$, and the corresponding approximation $\hat{\bx}_0^{t}$ (a
        proxy for the expectation of this distribution, i.e.,
        $\mathbb{E}_{\bx_0 \sim p_{\theta}(\cdot|\bx_t)}[\bx_0]$) for different values of $t$, together with
        the final generated image $\bx_0$.
    }
    \label{fig:diffusion_background}
\end{figure}

\section{A Data Attribution Framework for Diffusion Models}
\label{sec:prelim}
In this section, we introduce our framework for attributing samples generated by
diffusion models back to their training data.  To this end, we will specify both {\em what} to attribute as well
as how to {\em verify} the attributions. Specifically, in \Cref{sub:what} we
define data attribution for diffusion models as the task of understanding how
training data influence the {\em distribution} over the final images at each
step of the diffusion process.  Then, in \Cref{sub:eval}, we describe how to
evaluate and verify such attributions.

\subsection{Attributing the diffusion process step by step}
\label{sub:what}
Diffusion models generate images via a {\em multi-step} process. We thus decompose the task of attributing a final synthesized image into a corresponding series of
stages, with each stage providing attributions for a single step of the diffusion process. This stage-wise decomposition allows for:
\begin{itemize}
    \item \textbf{Fine-grained analysis.} Identifying influential training
    examples at each individual step gives us a fine-grained
    understanding of how data ``guides'' the diffusion process. This, in turn, allows us to capture, for example, that in some cases the same
    training example might be positively influential at early steps but
    negatively influential at later steps (see \cref{app:why-per-step}).
    \item \textbf{Computational feasibility.}
    Computing gradients through a single step requires only a single
    backwards pass. So, it becomes feasible to apply existing efficient data attribution
    methods~\citep{park2023trak,pruthi2020estimating} that involve computing
    gradients.
    \item \textbf{Feature-level attribution.} As we demonstrate below, features
    tend to form only within a small number of steps of the diffusion process.  Thus,
    attributing at an individual step level allows us to isolate influences of training points on formation of
    specific features within the final generated image.
\end{itemize}
It remains now to define \text{what} exactly to attribute to the training data
at each step. To this end, we first motivate studying the conditional
distribution $p_{\theta}(\cdot|\bx_t)$ (see \cref{sub:background_diffusion}) as
a way to quantify the impact of each step $t$ of the diffusion process to the
final sample $\bx_0$.  Next, we highlight how analyzing the evolution of this
distribution over steps $t$ can connect individual steps to specific features of
interest.  Finally, building on these observations, we formalize our framework
as attributing properties of this distribution $p_{\theta}(\cdot|\bx_t)$ at each
step $t$ to the training data.

\begin{figure}[t!]
  \centering
  \includegraphics[width=1\textwidth]{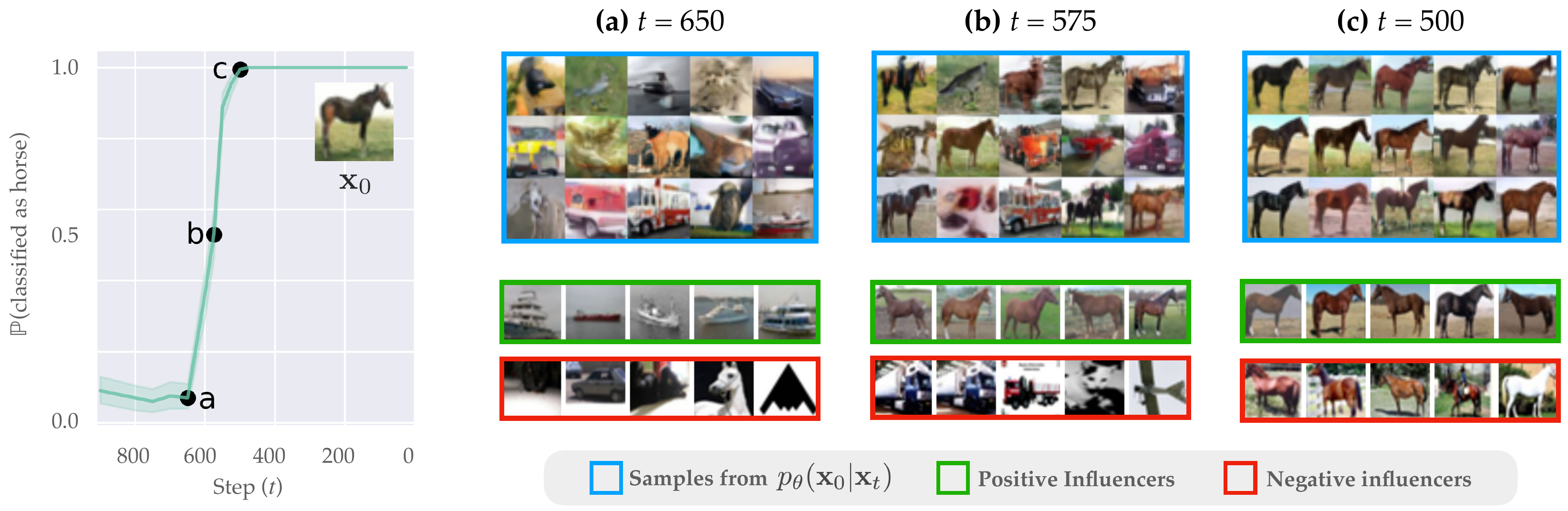}
  \caption{
      \textbf{Specific features appearing at specific steps.} \textbf{(Left)} For a given image of a horse ($\bx_0$)
      generated by a CIFAR-10 DDPM model, we plot the likelihood that samples
      from the distribution $p_{\theta}(\cdot|\bx_t)$ (see
      \cref{sub:background_diffusion}) are classified as a horse according to a
      CIFAR-10 classifier. This likelihood increases rapidly around steps $650$
      to $500$, suggesting that these steps are most responsible for the
      formation of this feature.
      (\textbf{Top}) For three steps $t$ in this range, we visualize samples
      from $p_{\theta}(\cdot|\bx_t)$.
      (\textbf{Bottom}) At each of these steps, we also visualize the training examples with the highest influence (positive in green, negative in red) identified by our method. Note that once the ``horse'' feature begins to appear (around $t=575$), positive influencers begin to reflect it. However, after this feature is ``decided'' (around $t=500$), negative influencers \textit{also} begin to reflect it.
  }
  \label{fig:phase_transition}
\end{figure}

\paragraph{Studying the distribution $p_{\theta}(\cdot|\bx_t)$.}
At a given step $t$ of the generative process, the relevant information
about the process up to that point is contained in the latent $\bx_t$.
While $\bx_t$ itself might not correspond to a natural image, we can use it to
directly sample from $p_{\theta}(\cdot|\bx_t)$, i.e., the distribution of possible
final images $\bx_0$ when resuming the diffusion process at step $t$ with
latent $\bx_t$. When $t=T$, this distribution is precisely the diffusion model's
learned distribution $p_{\theta}(\cdot)$, and at $t=0$ it is simply the final
sampled image $\bx_0$. So, intuitively, the progression of this conditional
distribution over steps $t$ informs us how the model gradually ``narrows down''
the possible distribution of samples to generate the final sample $\bx_0$ (see
\cref{fig:diffusion_background} for an illustration).  A natural way to
understand (and attribute) the impact of applying the diffusion process at each
step $t$ on the final image $\bx_0$ is thus to understand how this conditional
distribution $p_{\theta}(\cdot|\bx_t)$ evolves over steps.

\paragraph{Connecting features to specific steps.}
Given a final generated image, there might be many possible {\em features} of
interest within this image. For example, for $\bx_0$ in
\cref{fig:diffusion_background}, we might ask: {\em Why is there a grey bird?
Why is the background white?} How can we quantify the impact of a particular
step $t$ on a given feature in the final image? To answer this question, we
simply sample from the conditional distribution $p_{\theta}(\cdot|\bx_t)$ and
measure the fraction of samples that contain the feature of interest. Now, if we
treat this (empirical) likelihood as a function of $t$, the steps at which there
is the largest increase in (i.e., the steepest slope of) likelihood are most
responsible for the presence of this feature in the final image.

In fact, it turns out that such rapid increase in likelihood often happens
within only a small interval; we observe this phenomenon for both small-scale
unconditional models (DDPM trained on CIFAR-10, \cref{fig:phase_transition}) and
large-scale text-conditional models (Stable Diffusion v2 trained on LAION-5B,
\cref{app:features_stable}). As a result, we are able to tie the presence of a
given feature in the final image back to a small interval of steps $t$ in the
sampling process. In \cref{appfig:phase-classifier}, we further explore this
phenomenon for both different generated images and classifiers.

\paragraph{Implementing our approach.}
To implement our step-by-step attribution approach, we need a model output function
(see \cref{sub:background_diffusion}) that is specific to a step $t$. As we
motivated above, this function should be applied to samples from the conditional
distribution $p_{\theta}(\cdot | \bx_t)$.  To that end, we introduce a
step-specific model output function $f_t(p_{\theta(S)}(\cdot|\bx_t),
\theta(S))$. The function $f_t$ is intended to measure properties of the
distribution $p_{\theta(S)}(\cdot | \bx_t)$. For example, in \cref{sec:methods}
we define a concrete instantiation of $f_t$ that approximates the likelihood of
the model to generate individual samples from $p_{\theta(S)}(\cdot | \bx_t)$.
Adapting the general definition of data attribution from
\cref{sub:background_attribution}, we can now define {\em data attribution for
diffusion models} at a step $t$ as a function $\tau_t$ that assigns a score
$\tau_t(\bx_t,S)_i$ to each training example $z_i\in S$. This score indicates
the change in $f_t(p_{\theta(S)}(\cdot|\bx_t), \theta(S))$ induced by adding
$z_i$ to $S$

\subsection{Validating data attribution for diffusion models}
\label{sub:eval}

Visual inspection of the attributed training datapoints is a common heuristic
for evaluating the quality of data attribution. However, visual similarity is
not always reliable~\citep{ilyas2022datamodels, park2023trak}. In particular,
applications of data attribution such as data curation or model debugging often
require that the attributions are {\em causally predictive}.  Motivated by that,
we evaluate attribution scores according to how accurately they reflect the
corresponding training examples' {\em counterfactual} impact on the conditional
distribution $p_{\theta}(\cdot|\bx_t)$ using two different metrics. The first
metric (the linear datamodeling score) uses models trained on random subsets of
the full training set, whereas the second metric uses models trained on specific
counterfactual training sets targeted for each generated image.  The first
metric is cheaper to evaluate, as we can re-use the same set of models to
evaluate attributions for different target images and from different attribution
methods. On the other hand, the latter metric (retraining without the most
influential images) more directly measures changes in the conditional
distribution $p_{\theta}(\cdot|\bx_t)$, so we do not need to rely on a specific
choice of a model output function $f_t$.

\paragraph{Linear datamodeling score.}
The linear datamodeling score (LDS) is a measure of the effectiveness of a data attribution method that was introduced in \cite{ilyas2022datamodels,park2023trak} (see \cref{sub:background_attribution}). This metric quantifies how well the attribution scores can predict the exact {\em magnitude} of change in model output induced by (random) variations in the training set. In our setting, we apply it to the step-specific model output
function $f_t(p_{\theta(S)}(\cdot|\bx_t), \theta(S))$. Specifically, we use the
attribution scores $\tau$ to predict the diffusion-specific model output
function $f_t(p_{\theta(S)}(\cdot|\bx_t),\theta(S))$ as
\begin{equation}
  \label{eq:data_attr_agg}
  g_\tau(p_{\theta(S)}(\cdot|\bx_t), S'; S) \coloneqq \sum_{i\ :\ z_i \in S'} \tau(\bx_t, S)_i.
\end{equation}
Then, we can measure the
degree to which the predictions $g_\tau(p_{\theta(S)}(\cdot|\bx_t),S';S)$ are
correlated with the true outputs $f_t(p_{\theta(S)}(\cdot|\bx_t), \theta(S'))$ using
the LDS:
\[
  LDS(\tau, \bx_t) \coloneqq \bm{\rho}(\{f_t(p_{\theta(S)}(\cdot|\bx_t),\theta(S_j)): j \in [m]\},
  \{g_\tau(p_{\theta(S)}(\cdot|\bx_t), S_j;S): j \in [m]\}),
\]
where $\{S_1, \ldots, S_m: S_i \subset S\}$ are randomly sampled subsets of the
training set $S$ and $\bm{\rho}$ denotes Spearman's rank correlation
\citep{spearman1904proof}.  To decrease the cost of computing LDS, we use
$\hat{\bx}_0^{t}$ in lieu of samples from $p_{\theta(S)}(\cdot|\bx_t)$ (see
\cref{sub:background_diffusion}), since, as noted in
\cref{sub:background_diffusion}, $\hat{\bx}_0^{t}$ turns out to be a good proxy for the the latter quantity. In other words, we consider $f_t$ and $g_\tau$ as
functions of $\hat{\bx}_0^{t}$ rather than $p_{\theta(S)}(\cdot|\bx_t)$.

\paragraph{Retraining without the most influential images.}
In practice, we may want to use the data attributions to intentionally steer the
diffusion model's output. For example, we may want to remove all training
examples that cause the resulting model to generate a particular style of
images.  To evaluate the usefulness of a given data attribution method in these contexts, we
remove from the training set the most influential (i.e., highest scoring) images
for a given target  $\bx_t$, retrain a new model $\theta'$, then measure the
change in the conditional distribution $p_{\theta}(\cdot|\bx_t)$ (see
\cref{sub:background_diffusion}) when we replace $\theta$ with $\theta'$ only in
the neighborhood of step $t$ in the reverse diffusion process.  If the
data attributions are accurate, we expect the conditional distribution to change
significantly (as measured in our case using the FID distance for
images~\citep{heusel2017gans}).

As we consider data attributions that are specific to each step, in principle we
should use the denoising model {\em only} for the corresponding step $t$.
However, the impact of a single step on the final distribution might be small,
making it hard to measure.  Hence, we assume that attributions change only
gradually over steps and replace the denoising model for a {\em small interval}
of steps (i.e., between steps $t$ and $t-\Delta$).

\section{Efficiently Computing Attributions for Diffusion Models}
\label{sec:methods}
In this section, we describe how we can efficiently estimate data attributions for
diffusion models using \trak~\citep{park2023trak}.  As we described in
\cref{sub:background_attribution}, we can decompose the task of computing data
attribution scores into estimating two components: (i) the change in model
parameters, and (ii) the induced change in model output.  Following
\trak~\citep{park2023trak}, computing the first component (change in model
parameters) only requires computing per-example gradients of the training loss
(and in particular, does not require any re-training per each training datapoint). Similarly,
computing the second component (change in model output) only requires computing
gradients with respect to the model output function of choice (see Section 3 of
\citet{park2023trak} for details).  We now describe how to adapt the estimation
of the above two components to the diffusion model setting.

\paragraph{Estimating the change in model parameters.}
For diffusion models, the training process is much more complicated than the
standard supervised settings (e.g., image classification) considered in
\citet{park2023trak}. In particular, one challenge is that the diffusion model
outputs a high-dimensional vector (an image) as opposed to a single scalar
(e.g., a label). Even if we approximate the diffusion model as a {\em linear}
model in parameters, naively applying \trak would require keeping track of $p$
gradients for each training example (where $p$ is the number of pixels)
and thus be computationally infeasible. Nonetheless, it is still the case that
the presence of a single training example influences the optimization trajectory
{\em only} via the gradient of the loss on that example---specifically, the MSE
of the denoising objective. Hence, it suffices to keep track of a single
gradient for each example.  This observation allows us to estimate the change in
model parameters using the same approach that \trak uses (see
\cref{sub:background_attribution}).

An additional challenge is that the gradient updates in the diffusion process
are highly stochastic due to the sampling of random noise. To mitigate this
stochasticity, we average the training loss over multiple resampling of the
noise at randomly chosen steps and compute gradients over this averaged loss.

\begin{algorithm}[!h]
    \caption{\trak for diffusion models}
    \label{alg:dtrak}
    \begin{algorithmic}[1]
        \State {\bfseries Input:} %
        Model checkpoints $\{\thetastar_1, ..., \thetastar_M\}$,
        training dataset $S=\{\bz_1,...,\bz_N\}$,
        target sequence $\{\bx_1,...,\bx_T\}$ corresponding to $T$ steps,
        projection dimension $k \in \mathbb{N}$.
        \State {\bfseries Output:} Attribution scores $\tau(\bx_t, S) \in \mathbb{R}^{N}$ for each $t$
        \State $f_{\text{train}}(\bx,\theta) \coloneqq \mathbb{E}_{\bm{\varepsilon},t}
        \norm{\bm{\varepsilon} -
        \beps_{\theta(S)}\left(\sqrt{\bar\alpha_t}\bx + \sqrt{1 -
        \bar\alpha_t} \bm{\varepsilon},t\right)}_2^2$ \hfill$\triangleright$ DDPM training loss
        \State $f_t\left(\cdot, \theta\right)$ defined as in \Cref{eqn:scoring}
        \hfill$\triangleright$ Step-specific model output function $f_t(\cdot)$
        \For{$m \in \{1,\ldots,M\}$}
        \State $\mathbf{P} \sim \mathcal{N}(0, 1)^{p \times k}$ \hfill$\triangleright$ Sample random projection matrix
        \For{$i \in \{1,\ldots,N\}$}
        \State $\phi_i \gets \mathbf{P}^\top\nabla_\theta f_{\text{train}}(\bz_i, \thetastar_m)$ \hfill$\triangleright$ Compute training loss gradient at $\thetastar_m$ and project
        \EndFor
        \For{$t \in \{1,\ldots,T\}$}
        \State $\hat\bx^{(t)}_0 \gets c_1(\alpha_t)\cdot (\bx_t - c_2(\alpha_t \cdot \varepsilon_{\thetastar_m}(\bx_t,t)))$ \hfill$\triangleright$ Compute expectation of conditional distribution
        \State $g_i \gets \mathbf{P}^\top\nabla_\theta f_t(\hat\bx^{(t)}_0, \thetastar_m)$ \hfill$\triangleright$ Compute model output gradient at $\thetastar_m$ and project
        \EndFor
        \State $\Phi_m \gets [{\phi}_1; \cdots; {\phi}_N]^\top$
        \State $G_m \gets [g_1; \cdots; g_T]^\top$
        \EndFor
        \State
        $[\tau(\bx_1,S); \cdots; \tau(\bx_T,S)] \gets \frac{1}{m}\sum\limits_{m=1}^M
        \Phi_m(\Phi_m^\top \Phi_m)^{-1} G_m$  \hfill$\triangleright$ Average scores over checkpoints %
        \State \textbf{return} $\{\tau(\bx_t, S)\}$
    \end{algorithmic}
\end{algorithm}

\paragraph{A model output function for diffusion models.}
In \cref{sec:prelim}, we motivated why we would like to attribute properties of
the conditional distribution $p_{\theta(S)}(\cdot | \bx_t)$, i.e., the
distribution that arises from sampling when conditioning on an intermediate
latent $\bx_t$.  Specifically, we would like to understand what training data
causes the model to generate samples from this distribution.  Then, one natural
model output function $f_t$ would be to measure the likelihood of the model to
generate these samples. Attributing with respect to such a choice of $f_t$
allows us to understand what training examples increase or decrease this
likelihood.

In order to efficiently implement this model output function, we make two
simplifications.  First, sampling from $p_{\theta(S)}(\cdot | \bx_t)$ can be
computationally expensive, as this would involve repeatedly resampling parts of
the diffusion trajectory. Specifically, sampling once from $p_{\theta(S)}(\cdot
| \bx_t)$ requires applying the diffusion model $t$ times---in practice, $t$ can
often be as large as $1000$. Fortunately, as we described in
\cref{sub:background_diffusion}, we can use the one-step estimate
$\hat{\bx}_0^{t}$ as a proxy for samples from $p_{\theta(S)}(\cdot | \bx_t)$,
since it approximates this distribution's expectation $\mathbb{E}_{\bx_0 \sim
p_{\theta}(\cdot|\bx_t)}[\bx_0]$.

Second, it is computationally expensive to compute
gradients with respect to the exact likelihood of generating an image. So, as a
more tractable proxy for this likelihood, we measure the reconstruction
loss\footnote{ The reconstruction loss is a proxy for the likelihood of the
generated image, as it is proportional to the evidence lower bound
(ELBO)~\citep{sohldickstein2015deep,song2023consistency}.
}
(i.e., how well the diffusion model is able to denoise a noisy image) when
adding noise to $\hat{\bx}_0^{t}$ with magnitude matching the sampling process
at step $t$.  Specifically, we compute the Monte Carlo estimate
\begin{align}
    f_t\left(\hat{\bx}_0^{t}, \theta(S)\right) =
    \sum_{i=1}^k
    \norm{\bm{\varepsilon}_i -
    \beps_{\theta(S)}\left(\sqrt{\bar\alpha_t}\hat\bx^{(t)}_0 + \sqrt{1 -
    \bar\alpha_t} \bm{\varepsilon}_i,t\right)}_2^2,
    \label{eqn:scoring}
\end{align}
where $\bar\alpha_t$ is the DDPM\footnote{We only consider DDPM schedulers in
this work. The above derivation can be easily extended to other schedulers.}
variance schedule~\citep{ho2020denoising},
$\bm{\varepsilon}_i\sim\mathcal{N}(0,1)$ for all $i\in[k]$, and $k$ is the
number of resampling rounds of the random noise $\bm{\varepsilon}$.  Now that we
have chosen our model output function, we can simply compute gradients with
respect to this output to obtain the second component in \Cref{eq:attribution}.

    \paragraph{The final algorithm.}
    We summarize our algorithm for computing attribution scores in \Cref{alg:dtrak}.
    We approximate the training loss (line 3) with different samples of noise $\varepsilon$ and step $t$.
    Note that to attribute a new target sequence, we only have to recompute lines 10-12.

\section{Experiments}
\label{sec:experiments}
To evaluate our data attribution method, we apply it to DDPMs trained on
CIFAR-10 and LDMs trained on MS~COCO. First, in \cref{sec:attributions}, we
visually inspect and interpet our attributions, and then in
\cref{sec:evaluation} we evaluate their counterfactual significance using the
metrics we introduced in \cref{sub:eval}.  In \cref{sec:localizing}, we further
explore how our data attributions can be localized to patches in pixel space.
Finally, in \cref{sub:forgetting}, we investigate the value of our step-specific
attributions for attributing the full diffusion trajectory.

\subsection{Experimental setup}
We compute our data attribution scores using 100 DDPM checkpoints trained on CIFAR-10
and 50 LDM checkpoints trained MS~COCO (see Appendix~\ref{app:exp-details} for
training details.). As baselines, we compare our attributions to two common
image similarity metrics---CLIP similarity (i.e., cosine similarity in the CLIP
embedding space) and cosine similarity in pixel space.

\subsection{Qualitative analysis of attributions}
\label{sec:attributions}
In \cref{fig:hero_fig}, we visualize the sampling trajectory for an image
generated by an MS~COCO model, along with the most positive and negative
influencers identified by \trak (see \cref{app:omitted} for additional
visualizations of identified attributions on CIFAR-10 and \mscoco). We find that
positive influencers tend to resemble the generated image throughout, while
negative influencers tend to differ from the generated image along specific
attributes (e.g., class, background, color) depending on the step.
Interestingly, the negative influencers increasingly resemble the generated
image towards the end of the diffusion trajectory.

Intuitively, we might expect that negative influencers would not resemble the
final generated image, as they should to steer the trajectory away from that image. So, why
do they in fact reflect features of the final generated image?
To answer this question, we study the relationship between the top (positive and
negative) influencers and the distribution $p_{\theta}(\cdot|\bx_t)$ towards
which we target our attributions.
In \cref{fig:phase_transition}, for a given image of a horse generated by our
CIFAR-10 DDPM, we plot the likelihood that images from
$p_{\theta}(\cdot|\bx_t)$ containing a horse (according to a classifier trained on CIFAR-10) as a function of the step $t$ (left). We also show the top and
bottom influencers at three points along the trajectory (right). We find that
the top influencers begin to reflect the feature of interest once the likelihood
of this feature begins to grow. Yet, once the likelihood of the feature reaches
near certain, the negative influencers \textit{also} begin to reflect this
feature.  This behavior has the following intuitive explanation: after this
point, it would be impossible to ``steer'' the trajectory away from presenting
this feature. So, the negative influencers at later steps might now steer the
trajectory away from other features of the final image (e.g., the color of
horse) that has not yet been decided at that step. Additionally, images that do
not reflect the ``decided'' features might no longer be relevant to steering the
trajectory of the diffusion process.

\subsection{Counterfactually validating the attributions} \label{sec:evaluation}
We now evaluate our attributions using the metrics introduced in \cref{sub:eval} to validate their counterfactual significance.

\begin{figure}[!h]
    \centering
    \includegraphics[width=\linewidth]{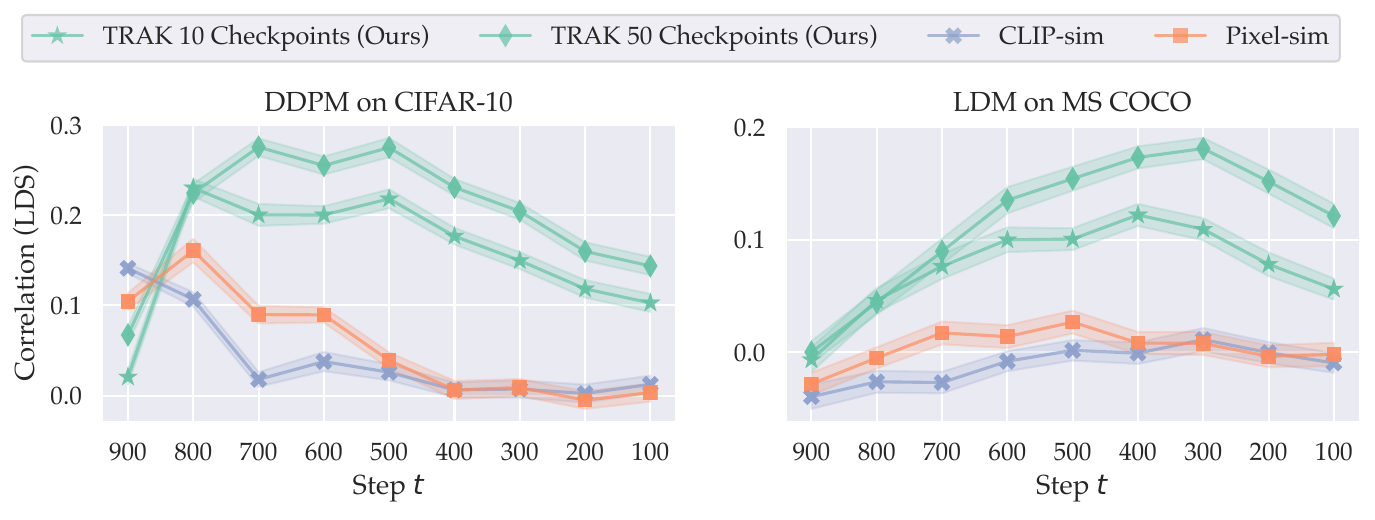}
    \vskip -.1cm
    \caption{\textbf{Predicting model behavior.} The counterfactual
    predictiveness of attributions measured using the LDS score along the diffusion trajectory (at every 100
    steps) for three different methods: \trak (computed using 10 and 50
    model checkpoints), CLIP similarity, and pixel similarity.
    Smaller steps are closer to the final sample.
    Shaded areas represent standard error.
    }
    \label{fig:lds}
    \end{figure}

\begin{figure}[!t]
    \centering
    \includegraphics[width=1\linewidth]{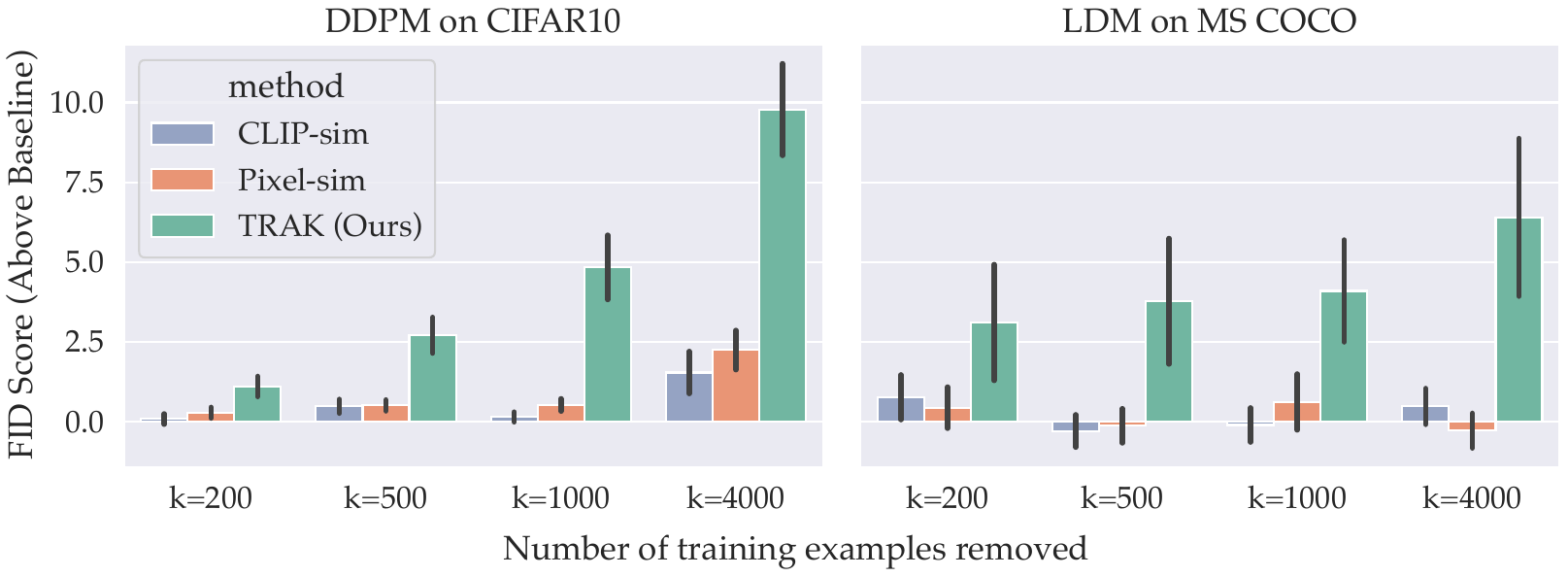}
    \vskip -.1cm
    \caption{\textbf{Retraining without top influencers.} Change
    in the distribution of generated images $p_{\theta}(\cdot|\bx_{400})$ when
    substituting the original model with a new model only between steps $400$
    and $300$. The new model is trained without the $k$ top influencers of
    $\bx_{400}$ according to attributions from \trak (computed at step
    $400$), CLIP similarity, or pixel similarity. To
    evaluate the change in distribution, we measure the increase in FID score
    over a baseline of models trained on the full dataset (see \Cref{sub:eval}
    for details). Bars represent standard error.}

    \label{fig:dist}
\end{figure}

\paragraph{LDS.} We sample 100 random $50\%$ subsets of CIFAR-10 and MS~COCO,
and train five models per mask. Given a set of attribution scores, we then compute
the Spearman rank correlation \citep{spearman1904proof} between the predicted
model outputs $g_\tau(\cdot)$ (see Eq.~(\ref{eq:data_attr_agg})) on each
training data subset according to the attributions and the (averaged) actual
model outputs.  To evaluate the counterfactual significance of our attributions
over the course of the diffusion trajectory, we measure LDS scores at every $100$
steps over the $1000$ step sampling process.

In \cref{fig:lds}, we plot LDS scores for CIFAR-10 (left) and MS~COCO (right)
over a range of steps for our attribution scores as well as the two similarity
baselines. Unlike in many computer vision
settings~\citep{zhang2018unreasonable}, we find that for CIFAR-10, similarity in
pixel space achieves competitive performance, especially towards the start of
the diffusion trajectory. However, for both CIFAR-10 and MS~COCO, only \trak is
counterfactually predictive across the entire trajectory.

\paragraph{Retraining without the most influential images.}

We compute
attribution scores on 50 samples from our CIFAR-10 and MS~COCO models
at step $t = 400$.  Given the attribution scores for each sample, we then
retrain the model after removing the corresponding top $k$ influencers for
$k\in\{200,500,1000\}$. We sample $5000$ images from two distributions: (1) the
distribution arising from repeatedly initializing at $\bx_{400}$ and sampling
the final 400 steps from the original model; and (2) the distribution arising
from repeating the above process but using the retrained model only for steps
$t=400$ to $t=300$.  We then compute FID distance between these distributions,
and repeat this process for each sample at each value of $k$.

In \cref{fig:dist}, we display the average FID scores (a measure of distance
from the original model) after removing the $k$ most influential images for a
given sample across possible values of $k$. We notice that, for all values of $k$,
removing the top influencers identified by our attribution method has a greater
impact than removing the most similar images according to CLIP or pixel space
similarities.

\begin{figure*}[!h]
    \centering
    \includegraphics[width=.98\linewidth]{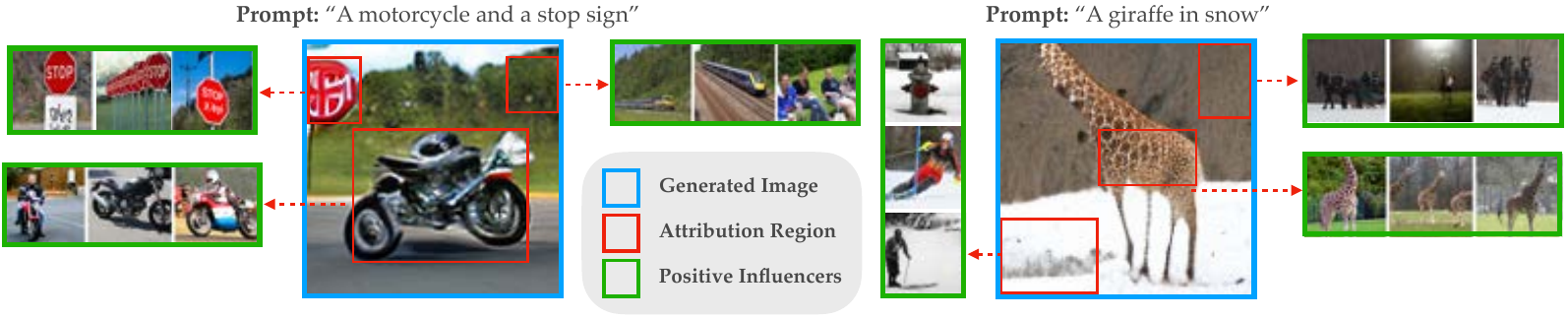}
    \caption{\textbf{Patch-based attribution.}
     We adapt our method to restrict attribution to user-specified patches of a
     generated image. We show examples of attributing patches capturing
     individual concepts in images synthesized by a latent diffusion model
     trained on MS~COCO. Attributions are computed at step $t=400$.}
    \label{fig:attributing-patches}
\end{figure*}

\subsection{Localizing our attributions to patches in pixel space}
\label{sec:localizing}
In \Cref{sec:prelim}, we discussed how step-by-step attribution allows us to
attribute particular features appearing within a particular interval of steps.
However, some features may appear together within a small interval, making it
hard to isolate them only based on the step.  Here we explore one possible
approach for better isolating individual features: selecting a region of pixels
(i.e., a \textit{patch}) in a generated sample corresponding to a feature of
interest, and restricting our model output function to this region. This way, we
can restrict attributions only to the selected patch, which can be useful for
understanding what caused a specific feature to appear (see
\cref{fig:attributing-patches}). To implement this model output function, we
simply apply a pixel-wise binary mask to \cref{eqn:scoring} and ignore the
output outside of the masked region.  To test this approach, we generate images
containing multiple features with an MS~COCO-trained LDM. We then manually
create per-feature masks for which we compute attribution scores with our method
(see Figure~\ref{fig:attributing-patches}). The resulting attributions for
different masks surface training examples relevant \textit{only} to the
corresponding features in that region.

\begin{figure}[!t]
    \centering
    \includegraphics[width=0.37\linewidth]{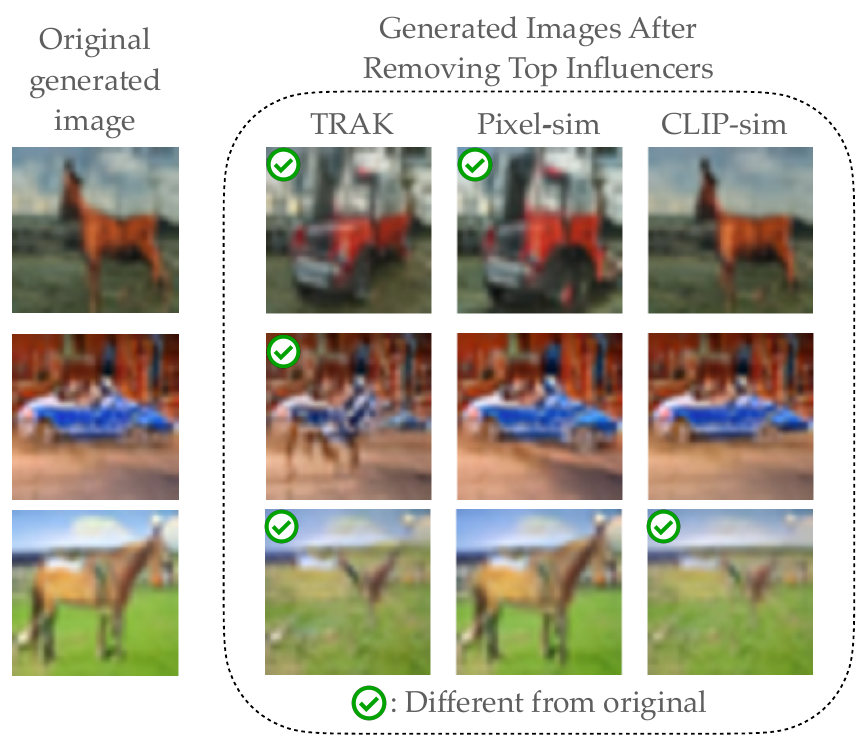}
    \hfill
    \includegraphics[width=0.59\linewidth]{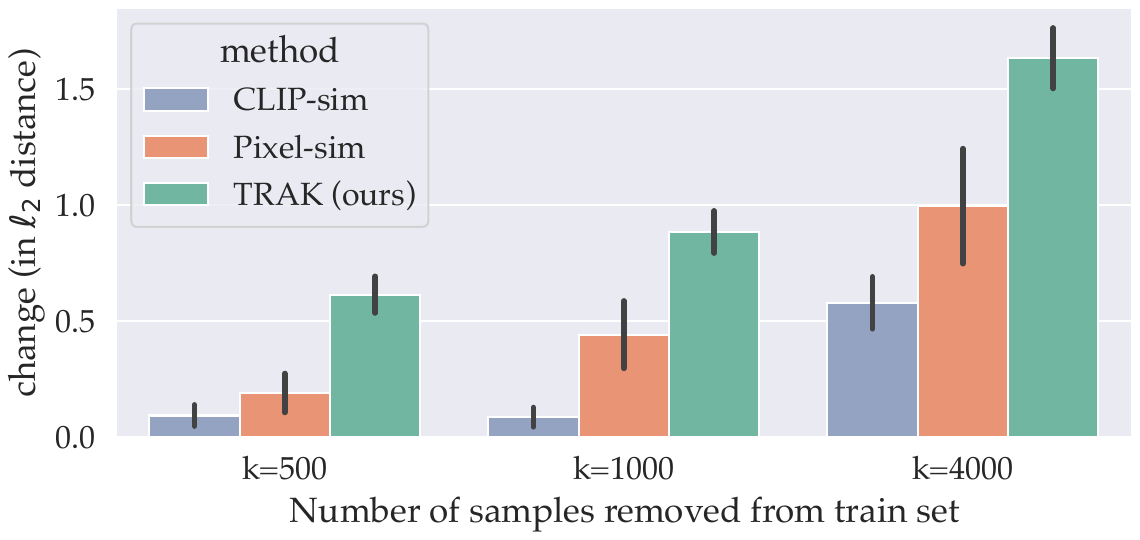}
    \vskip -.1cm
    \caption{\textbf{``Forgetting'' an image.} We quantify the
    impact of removing the highest scoring training examples according to \trak,
    CLIP similarity, and pixel similarity (and re-training).  \textbf{(Left)} We
    compare the original synthesized samples to those generated from the same
    random seed with the re-trained models.  \textbf{(Right)} To quantify the
    impact of removing these images, we measure the $\ell_2$ distance between 60
    synthesized samples and corresponding images generated by the re-trained
    models. Black bars represent standard error.
    }
    \label{fig:counterfactual}
\end{figure}

\subsection{``Forgetting'' how to generate an image}
\label{sub:forgetting}
Our attribution scores and evaluation metrics are all step-specific. However, in practice we might care about identifying training images that impact the \textit{full} diffusion pipeline. In particular, we might be interested in whether removing the important training images for a given synthesized image causes the diffusion model to ``forget'' how to generate this image.

Specifically, given a set of attribution scores for a synthesized image, we
remove the top $k$ influencers (at step $t=300$), retrain the model, and generate new images from
scratch using the same random seed. Here, we leverage the fact that two
diffusion models trained on the same dataset tend to generate similar images
given the same random seed (see \cref{app:seed-consistency} for more details).
We then compare the change in pixel space between the original and newly
generated image. This process is distinct from our second evaluation metric, as (1) we
directly compare two images rather than measure the distance between
distributions, and (2) we re-generate images with our new model from scratch
rather than restarting from some intermediate latent $\bx_t$ and substituting
the new model for only a small interval of steps (between $t$ and $t-\Delta$).

We perform this process for our attribution scores on CIFAR-10 as well as the
two similarity baselines (see \cref{fig:counterfactual}). Our results
suggest that \trak is able to identify influential images that have a
significant impact on the full diffusion trajectory of the diffusion model.

\section{Related Work}
\label{sec:rel_work}
\paragraph*{Data attribution.}
A long line of work has studied the problem of training data attribution, or tracing model behavior back to training data; we focus here on works done in the context of modern machine learning algorithms.
Prior approaches include those based on the influence function and its variants \citep{hampel2011robust,wojnowicz2016influence,koh2017understanding,basu2019second,khanna2019interpreting,achille2021lqf,schioppa2022scaling,bae2022if},
sampling-based methods that leverage models trained on different subsets of data \citep{ghorbani2019data,jia2019towards,feldman2020neural,ilyas2022datamodels,lin2022measuring}, and various other heuristic approaches \citep{yeh2018representer,pruthi2020estimating}.
These methods generally exhibit a strong tradeoff between predictiveness or effectiveness and computational efficiency \cite{jia2021scalability}. The recent method of \citet{park2023trak} significantly improves upon these tradeoffs by leveraging the empirical kernel structure of differentiable models.
While most prior work primarily focus on the supervised setting, more recent works study attribution in generative settings, including to language models \citep{park2023trak} and to diffusion models \citep{wang2023evaluating}.

In a recent work, \citet{wang2023evaluating} propose a method for {\em
efficiently evaluating} data attribution methods for generative models by
creating custom datasets with known ground-truth attributions.

\paragraph*{Memorization in generative models.}
We can view \textit{memorization} as a special case of data attribution where only few, nearly identical images
in the training set are responsible for the generation of a corresponding image.
Prior to the increasing
popularity of diffusion models, a number of previous works studied memorization
in other generative models. For example, \citet{feng2021when} study the impact
of properties of a dataset (size, complexity) on training data replication in
Generative Adversarial Networks (GANs), and \citet{burg2021memorization}
introduce a memorization score for Variational Autoencoders (VAEs) that can be
additionally applied to arbitrary generative models. Following the release of
large text-to-image diffusion models, the creators of one of these models
(DALL·E 2) investigated memorization issues themselves and found that
memorization could be significantly decreased through de-duplication of the
training data \citep{nichol2022dalle}. Recently, \citet{somepalli2022diffusion}
explore the data replication behavior of diffusion models from the lens of
``digital forgery,'' and identify many cases where, even when Stable Diffusion
produces ``unique'' images, it directly copies style and semantic structure from
individual images in the training set. On the other hand,
\citet{carlini2023extracting} investigate memorization from the perspective of
privacy, and show that query access to diffusion models can enable an adversary
to directly extract the models' training data.

\section{Conclusion}
\label{sec:conclusion}
In this work, we introduce a framework for data attribution for diffusion models
and provide an efficient method for computing such attributions. In particular,
we formalize data attribution in this setting as task of
quantifying how individual training datapoints influences the distribution over final images \textit{at each step} of the diffusion process. We demonstrate the efficacy of our approach on
DDPMs trained on CIFAR-10 and LDMs trained on
MS~COCO. Our framework also constitutes a step towards better understanding of how training data influences diffusion models.

There are several directions for potential improvements and future work. First,
our particular instantiation of the framework relies on proxies for the distribution
$p_\theta(\cdot|x_t)$ of final generated images conditioned on a given step $t$,
as well as for the likelihood of generating a given image. So, identifying more
accurate proxies could help improve the quality of the resulting attributions.
More broadly, we evaluate our framework on two academic-size datasets,
but the most popular diffusion models (such as Stable Diffusion) are
 larger and trained on significantly larger datasets. Thus, while feasible in principle, scaling our framework to such settings is important. Finally, while we study
the task of attributing individual steps, it would be valuable to perform data
attribution for the full diffusion process.

\section*{Acknowledgements}
The authors would like to thank Andrew Ilyas and Giannis Daras for helpful discussions.

Work supported in part by the NSF grants CNS-1815221 and DMS-2134108, and Open Philanthropy. This material is based upon work supported by the Defense Advanced Research Projects Agency (DARPA) under Contract No. HR001120C0015.

Research was sponsored by the United States Air Force Research Laboratory and the Department of the Air Force Artificial Intelligence Accelerator and was accomplished under Cooperative Agreement Number FA8750-19-2-1000. The views and conclusions contained in this document are those of the authors and should not be interpreted as representing the official policies, either expressed or implied, of the Department of the Air Force or the U.S. Government. The U.S. Government is authorized to reproduce and distribute reprints for Government purposes notwithstanding any copyright notation herein.

\bibliography{bibliography/fmt_bib.bib}

\begin{thebibliography}{54}
\providecommand{\natexlab}[1]{#1}
\providecommand{\url}[1]{\texttt{#1}}
\expandafter\ifx\csname urlstyle\endcsname\relax
  \providecommand{\doi}[1]{doi: #1}\else
  \providecommand{\doi}{doi: \begingroup \urlstyle{rm}\Url}\fi

\bibitem[Ramesh et~al.(2022)Ramesh, Dhariwal, Nichol, Chu, and
  Chen]{ramesh2022hierarchical}
Aditya Ramesh, Prafulla Dhariwal, Alex Nichol, Casey Chu, and Mark Chen.
\newblock Hierarchical text-conditional image generation with clip latents.
\newblock \emph{arXiv preprint arXiv:2204.06125}, 2022.

\bibitem[Rombach et~al.(2022)Rombach, Blattmann, Lorenz, Esser, and
  Ommer]{rombach2022high}
Robin Rombach, Andreas Blattmann, Dominik Lorenz, Patrick Esser, and Bj{\"o}rn
  Ommer.
\newblock High-resolution image synthesis with latent diffusion models.
\newblock In \emph{Proceedings of the IEEE/CVF Conference on Computer Vision
  and Pattern Recognition}, pages 10684--10695, 2022.

\bibitem[Schuhmann et~al.(2022)Schuhmann, Beaumont, Vencu, Gordon, Wightman,
  Cherti, Coombes, Katta, Mullis, Wortsman, et~al.]{schuhmann2022laion}
Christoph Schuhmann, Romain Beaumont, Richard Vencu, Cade Gordon, Ross
  Wightman, Mehdi Cherti, Theo Coombes, Aarush Katta, Clayton Mullis, Mitchell
  Wortsman, et~al.
\newblock Laion-5b: An open large-scale dataset for training next generation
  image-text models.
\newblock In \emph{arXiv preprint arXiv:2210.08402}, 2022.

\bibitem[Carlini et~al.(2021)Carlini, Tramer, Wallace, Jagielski, Herbert-Voss,
  Lee, Roberts, Brown, Song, Erlingsson, et~al.]{carlini2021extracting}
Nicholas Carlini, Florian Tramer, Eric Wallace, Matthew Jagielski, Ariel
  Herbert-Voss, Katherine Lee, Adam Roberts, Tom Brown, Dawn Song, Ulfar
  Erlingsson, et~al.
\newblock Extracting training data from large language models.
\newblock In \emph{30th {USENIX} Security Symposium ({USENIX} Security 21)},
  2021.

\bibitem[Somepalli et~al.(2022)Somepalli, Singla, Goldblum, Geiping, and
  Goldstein]{somepalli2022diffusion}
Gowthami Somepalli, Vasu Singla, Micah Goldblum, Jonas Geiping, and Tom
  Goldstein.
\newblock Diffusion art or digital forgery? investigating data replication in
  diffusion models.
\newblock \emph{arXiv preprint arXiv:2212.03860}, 2022.

\bibitem[Shah et~al.(2022)Shah, Park, Ilyas, and Madry]{shah2022modeldiff}
Harshay Shah, Sung~Min Park, Andrew Ilyas, and Aleksander Madry.
\newblock Modeldiff: A framework for comparing learning algorithms.
\newblock In \emph{arXiv preprint arXiv:2211.12491}, 2022.

\bibitem[Lin et~al.(2022)Lin, Zhang, Lecuyer, Li, Panda, and
  Sen]{lin2022measuring}
Jinkun Lin, Anqi Zhang, Mathias Lecuyer, Jinyang Li, Aurojit Panda, and
  Siddhartha Sen.
\newblock Measuring the effect of training data on deep learning predictions
  via randomized experiments.
\newblock \emph{arXiv preprint arXiv:2206.10013}, 2022.

\bibitem[Khanna et~al.(2019)Khanna, Kim, Ghosh, and
  Koyejo]{khanna2019interpreting}
Rajiv Khanna, Been Kim, Joydeep Ghosh, and Sanmi Koyejo.
\newblock Interpreting black box predictions using fisher kernels.
\newblock In \emph{The 22nd International Conference on Artificial Intelligence
  and Statistics}, 2019.

\bibitem[Andersen et~al.(2023)Andersen, McKernan, and Ortiz]{andersen2023class}
Sarah Andersen, Kelly McKernan, and Karla Ortiz.
\newblock Class-action com­plaint against stability ai, 2023.
\newblock URL
  \url{https://stablediffusionlitigation.com/pdf/00201/1-1-stable-diffusion-complaint.pdf}.
\newblock Case 3:23-cv-00201.

\bibitem[Images(2023)]{images2023getty}
Getty Images.
\newblock Getty images (us), inc. v. stability ai, inc, 2023.
\newblock URL
  \url{https://fingfx.thomsonreuters.com/gfx/legaldocs/byvrlkmwnve/GETTY%20IMAGES%20AI%20LAWSUIT%20complaint.pdf}.
\newblock Case 1:23-cv-00135-UNA.

\bibitem[Azizi et~al.(2023)Azizi, Kornblith, Saharia, Norouzi, and
  Fleet]{azizi2023synthetic}
Shekoofeh Azizi, Simon Kornblith, Chitwan Saharia, Mohammad Norouzi, and
  David~J Fleet.
\newblock Synthetic data from diffusion models improves imagenet
  classification.
\newblock \emph{arXiv preprint arXiv:2304.08466}, 2023.

\bibitem[Kattakinda et~al.(2022)Kattakinda, Levine, and
  Feizi]{kattakinda2022invariant}
Priyatham Kattakinda, Alexander Levine, and Soheil Feizi.
\newblock Invariant learning via diffusion dreamed distribution shifts.
\newblock \emph{arXiv preprint arXiv:2211.10370}, 2022.

\bibitem[Wiles et~al.(2022)Wiles, Albuquerque, and Gowal]{wiles2022discovering}
Olivia Wiles, Isabela Albuquerque, and Sven Gowal.
\newblock Discovering bugs in vision models using off-the-shelf image
  generation and captioning.
\newblock \emph{arXiv preprint arXiv:2208.08831}, 2022.

\bibitem[Vendrow et~al.(2023)Vendrow, Jain, Engstrom, and
  Madry]{vendrow2023dataset}
Joshua Vendrow, Saachi Jain, Logan Engstrom, and Aleksander Madry.
\newblock Dataset interfaces: Diagnosing model failures using controllable
  counterfactual generation.
\newblock \emph{arXiv preprint arXiv:2302.07865}, 2023.

\bibitem[Luccioni et~al.(2023)Luccioni, Akiki, Mitchell, and
  Jernite]{luccioni2023stable}
Alexandra~Sasha Luccioni, Christopher Akiki, Margaret Mitchell, and Yacine
  Jernite.
\newblock Stable bias: Analyzing societal representations in diffusion models.
\newblock In \emph{arXiv preprint arXiv:2303.11408}, 2023.

\bibitem[Perera and Patel(2023)]{perera2023analyzing}
Malsha~V Perera and Vishal~M Patel.
\newblock Analyzing bias in diffusion-based face generation models.
\newblock In \emph{arXiv preprint arXiv:2305.06402}, 2023.

\bibitem[Koh and Liang(2017)]{koh2017understanding}
Pang~Wei Koh and Percy Liang.
\newblock Understanding black-box predictions via influence functions.
\newblock In \emph{International Conference on Machine Learning}, 2017.

\bibitem[Ghorbani et~al.(2019)Ghorbani, Wexler, Zou, and
  Kim]{ghorbani2019towards}
Amirata Ghorbani, James Wexler, James Zou, and Been Kim.
\newblock Towards automatic concept-based explanations.
\newblock \emph{arXiv preprint arXiv:1902.03129}, 2019.

\bibitem[Jia et~al.(2019)Jia, Dao, Wang, Hubis, Hynes, G\"{u}rel, Li, Zhang,
  Song, and Spanos]{jia2019towards}
Ruoxi Jia, David Dao, Boxin Wang, Frances~Ann Hubis, Nick Hynes, Nezihe~Merve
  G\"{u}rel, Bo~Li, Ce~Zhang, Dawn Song, and Costas~J. Spanos.
\newblock Towards efficient data valuation based on the shapley value.
\newblock In \emph{Proceedings of the Twenty-Second International Conference on
  Artificial Intelligence and Statistics}, 2019.

\bibitem[Ilyas et~al.(2022)Ilyas, Park, Engstrom, Leclerc, and
  Madry]{ilyas2022datamodels}
Andrew Ilyas, Sung~Min Park, Logan Engstrom, Guillaume Leclerc, and Aleksander
  Madry.
\newblock Datamodels: Predicting predictions from training data.
\newblock In \emph{International Conference on Machine Learning (ICML)}, 2022.

\bibitem[Hammoudeh and Lowd(2022)]{hammoudeh2022training}
Zayd Hammoudeh and Daniel Lowd.
\newblock Training data influence analysis and estimation: A survey.
\newblock In \emph{arXiv preprint arXiv:2212.04612}, 2022.

\bibitem[Park et~al.(2023)Park, Georgiev, Ilyas, Leclerc, and
  Madry]{park2023trak}
Sung~Min Park, Kristian Georgiev, Andrew Ilyas, Guillaume Leclerc, and
  Aleksander Madry.
\newblock Trak: Attributing model behavior at scale.
\newblock In \emph{Arxiv preprint arXiv:2303.14186}, 2023.

\bibitem[Ho et~al.(2020)Ho, Jain, and Abbeel]{ho2020denoising}
Jonathan Ho, Ajay Jain, and Pieter Abbeel.
\newblock Denoising diffusion probabilistic models.
\newblock In \emph{Neural Information Processing Systems (NeurIPS)}, 2020.

\bibitem[Krizhevsky(2009)]{krizhevsky2009learning}
Alex Krizhevsky.
\newblock Learning multiple layers of features from tiny images.
\newblock In \emph{Technical report}, 2009.

\bibitem[Lin et~al.(2014)Lin, Maire, Belongie, Hays, Perona, Ramanan,
  Doll{\'a}r, and Zitnick]{lin2014microsoft}
Tsung-Yi Lin, Michael Maire, Serge Belongie, James Hays, Pietro Perona, Deva
  Ramanan, Piotr Doll{\'a}r, and C~Lawrence Zitnick.
\newblock Microsoft coco: Common objects in context.
\newblock In \emph{European conference on computer vision (ECCV)}, 2014.

\bibitem[Wojnowicz et~al.(2016)Wojnowicz, Cruz, Zhao, Wallace, Wolff, Luan, and
  Crable]{wojnowicz2016influence}
Mike Wojnowicz, Ben Cruz, Xuan Zhao, Brian Wallace, Matt Wolff, Jay Luan, and
  Caleb Crable.
\newblock Influence sketching: Finding influential samples in large-scale
  regressions.
\newblock In \emph{2016 IEEE International Conference on Big Data (Big Data)},
  2016.

\bibitem[Pruthi et~al.(2020)Pruthi, Liu, Sundararajan, and
  Kale]{pruthi2020estimating}
Garima Pruthi, Frederick Liu, Mukund Sundararajan, and Satyen Kale.
\newblock Estimating training data influence by tracing gradient descent.
\newblock In \emph{Neural Information Processing Systems (NeurIPS)}, 2020.

\bibitem[Long(2021)]{long2021properties}
Philip~M Long.
\newblock Properties of the after kernel.
\newblock In \emph{arXiv preprint arXiv:2105.10585}, 2021.

\bibitem[Wei et~al.(2022)Wei, Hu, and Steinhardt]{wei2022more}
Alexander Wei, Wei Hu, and Jacob Steinhardt.
\newblock More than a toy: Random matrix models predict how real-world neural
  representations generalize.
\newblock In \emph{ICML}, 2022.

\bibitem[Malladi et~al.(2022)Malladi, Wettig, Yu, Chen, and
  Arora]{malladi2022kernel}
Sadhika Malladi, Alexander Wettig, Dingli Yu, Danqi Chen, and Sanjeev Arora.
\newblock A kernel-based view of language model fine-tuning.
\newblock In \emph{arXiv preprint arXiv:2210.05643}, 2022.

\bibitem[Sohl-Dickstein et~al.(2015)Sohl-Dickstein, Weiss, Maheswaranathan, and
  Ganguli]{sohldickstein2015deep}
Jascha Sohl-Dickstein, Eric~A. Weiss, Niru Maheswaranathan, and Surya Ganguli.
\newblock Deep unsupervised learning using nonequilibrium thermodynamics.
\newblock In \emph{International Conference on Machine Learning}, 2015.

\bibitem[Song and Ermon(2019)]{song2019generative}
Yang Song and Stefano Ermon.
\newblock Generative modeling by estimating gradients of the data distribution.
\newblock In \emph{Neural Information Processing Systems (NeurIPS)}, 2019.

\bibitem[Song et~al.(2023)Song, Dhariwal, Chen, and
  Sutskever]{song2023consistency}
Yang Song, Prafulla Dhariwal, Mark Chen, and Ilya Sutskever.
\newblock Consistency models.
\newblock \emph{arXiv preprint arXiv:2303.01469}, 2023.

\bibitem[Daras et~al.(2023)Daras, Dagan, Dimakis, and
  Daskalakis]{daras2023consistent}
Giannis Daras, Yuval Dagan, Alexandros~G Dimakis, and Constantinos Daskalakis.
\newblock Consistent diffusion models: Mitigating sampling drift by learning to
  be consistent.
\newblock \emph{arXiv preprint arXiv:2302.09057}, 2023.

\bibitem[Song et~al.(2021)Song, Sohl-Dickstein, Kingma, Kumar, Ermon, and
  Poole]{song2021score}
Yang Song, Jascha Sohl-Dickstein, Diederik~P Kingma, Abhishek Kumar, Stefano
  Ermon, and Ben Poole.
\newblock Score-based generative modeling through stochastic differential
  equations.
\newblock In \emph{International Conference on Learning Representations}, 2021.
\newblock URL \url{https://openreview.net/forum?id=PxTIG12RRHS}.

\bibitem[Ho and Salimans(2022)]{ho2022classifier}
Jonathan Ho and Tim Salimans.
\newblock Classifier-free diffusion guidance.
\newblock \emph{arXiv preprint arXiv:2207.12598}, 2022.

\bibitem[Spearman(1904)]{spearman1904proof}
Charles Spearman.
\newblock The proof and measurement of association between two things.
\newblock In \emph{The American Journal of Psychology}, 1904.

\bibitem[Heusel et~al.(2017)Heusel, Ramsauer, Unterthiner, Nessler, and
  Hochreiter]{heusel2017gans}
Martin Heusel, Hubert Ramsauer, Thomas Unterthiner, Bernhard Nessler, and Sepp
  Hochreiter.
\newblock Gans trained by a two time-scale update rule converge to a local nash
  equilibrium.
\newblock In \emph{Neural Information Processing Systems (NeurIPS)}, 2017.

\bibitem[Zhang et~al.(2018)Zhang, Isola, Efros, Shechtman, and
  Wang]{zhang2018unreasonable}
Richard Zhang, Phillip Isola, Alexei~A Efros, Eli Shechtman, and Oliver Wang.
\newblock The unreasonable effectiveness of deep features as a perceptual
  metric.
\newblock In \emph{Computer Vision and Pattern Recognition (CVPR)}, 2018.

\bibitem[Hampel et~al.(2011)Hampel, Ronchetti, Rousseeuw, and
  Stahel]{hampel2011robust}
Frank~R Hampel, Elvezio~M Ronchetti, Peter~J Rousseeuw, and Werner~A Stahel.
\newblock \emph{Robust statistics: the approach based on influence functions},
  volume 196.
\newblock John Wiley \& Sons, 2011.

\bibitem[Basu et~al.(2019)Basu, You, and Feizi]{basu2019second}
Samyadeep Basu, Xuchen You, and Soheil Feizi.
\newblock Second-order group influence functions for black-box predictions.
\newblock In \emph{International Conference on Machine Learning (ICML)}, 2019.

\bibitem[Achille et~al.(2021)Achille, Golatkar, Ravichandran, Polito, and
  Soatto]{achille2021lqf}
Alessandro Achille, Aditya Golatkar, Avinash Ravichandran, Marzia Polito, and
  Stefano Soatto.
\newblock Lqf: Linear quadratic fine-tuning.
\newblock In \emph{Proceedings of the IEEE/CVF Conference on Computer Vision
  and Pattern Recognition}, 2021.

\bibitem[Schioppa et~al.(2022)Schioppa, Zablotskaia, Vilar, and
  Sokolov]{schioppa2022scaling}
Andrea Schioppa, Polina Zablotskaia, David Vilar, and Artem Sokolov.
\newblock Scaling up influence functions.
\newblock In \emph{Proceedings of the AAAI Conference on Artificial
  Intelligence}, volume~36, pages 8179--8186, 2022.

\bibitem[Bae et~al.(2022)Bae, Ng, Lo, Ghassemi, and Grosse]{bae2022if}
Juhan Bae, Nathan Ng, Alston Lo, Marzyeh Ghassemi, and Roger Grosse.
\newblock If influence functions are the answer, then what is the question?
\newblock In \emph{ArXiv preprint arXiv:2209.05364}, 2022.

\bibitem[Ghorbani and Zou(2019)]{ghorbani2019data}
Amirata Ghorbani and James Zou.
\newblock Data shapley: Equitable valuation of data for machine learning.
\newblock In \emph{International Conference on Machine Learning (ICML)}, 2019.

\bibitem[Feldman and Zhang(2020)]{feldman2020neural}
Vitaly Feldman and Chiyuan Zhang.
\newblock What neural networks memorize and why: Discovering the long tail via
  influence estimation.
\newblock In \emph{Advances in Neural Information Processing Systems
  (NeurIPS)}, volume~33, pages 2881--2891, 2020.

\bibitem[Yeh et~al.(2018)Yeh, Kim, Yen, and Ravikumar]{yeh2018representer}
Chih-Kuan Yeh, Joon~Sik Kim, Ian E.~H. Yen, and Pradeep Ravikumar.
\newblock Representer point selection for explaining deep neural networks.
\newblock In \emph{Neural Information Processing Systems (NeurIPS)}, 2018.

\bibitem[Jia et~al.(2021)Jia, Wu, Sun, Xu, Dao, Kailkhura, Zhang, Li, and
  Song]{jia2021scalability}
Ruoxi Jia, Fan Wu, Xuehui Sun, Jiacen Xu, David Dao, Bhavya Kailkhura,
  Ce~Zhang, Bo~Li, and Dawn Song.
\newblock Scalability vs. utility: Do we have to sacrifice one for the other in
  data importance quantification?
\newblock In \emph{Proceedings of the IEEE/CVF Conference on Computer Vision
  and Pattern Recognition}, 2021.

\bibitem[Wang et~al.(2023)Wang, Efros, Zhu, and Zhang]{wang2023evaluating}
Sheng-Yu Wang, Alexei~A Efros, Jun-Yan Zhu, and Richard Zhang.
\newblock Evaluating data attribution for text-to-image models.
\newblock \emph{arXiv preprint arXiv:2306.09345}, 2023.

\bibitem[Feng et~al.(2021)Feng, Guo, Benitez-Quiroz, and
  Martinez]{feng2021when}
Qianli Feng, Chenqi Guo, Fabian Benitez-Quiroz, and Aleix~M Martinez.
\newblock When do gans replicate? on the choice of dataset size.
\newblock In \emph{Proceedings of the IEEE/CVF International Conference on
  Computer Vision}, pages 6701--6710, 2021.

\bibitem[van~den Burg and Williams(2021)]{burg2021memorization}
Gerrit van~den Burg and Chris Williams.
\newblock On memorization in probabilistic deep generative models.
\newblock \emph{Advances in Neural Information Processing Systems},
  34:\penalty0 27916--27928, 2021.

\bibitem[Nichol et~al.(2022)Nichol, Ramesh, Mishkin, Dariwal, Jang, and
  Chen]{nichol2022dalle}
Alex Nichol, Aditya Ramesh, Pamela Mishkin, Prafulla Dariwal, Joanne Jang, and
  Mark Chen.
\newblock Dalle 2 pre-training mitigations.
\newblock 2022.

\bibitem[Carlini et~al.(2023)Carlini, Hayes, Nasr, Jagielski, Sehwag, Tramer,
  Balle, Ippolito, and Wallace]{carlini2023extracting}
Nicholas Carlini, Jamie Hayes, Milad Nasr, Matthew Jagielski, Vikash Sehwag,
  Florian Tramer, Borja Balle, Daphne Ippolito, and Eric Wallace.
\newblock Extracting training data from diffusion models.
\newblock \emph{arXiv preprint arXiv:2301.13188}, 2023.

\bibitem[Radford et~al.(2021)Radford, Kim, Hallacy, Ramesh, Goh, Agarwal,
  Sastry, Askell, Mishkin, Clark, et~al.]{radford2021learning}
Alec Radford, Jong~Wook Kim, Chris Hallacy, Aditya Ramesh, Gabriel Goh,
  Sandhini Agarwal, Girish Sastry, Amanda Askell, Pamela Mishkin, Jack Clark,
  et~al.
\newblock Learning transferable visual models from natural language
  supervision.
\newblock In \emph{arXiv preprint arXiv:2103.00020}, 2021.

\end{thebibliography}

\clearpage
\appendix
\addcontentsline{toc}{section}{Appendix}
\renewcommand\ptctitle{Appendices}
\part{}
\parttoc
\clearpage

\counterwithin{figure}{section}
\counterwithin{table}{section}
\counterwithin{algorithm}{section}

\newpage
\appendix
\onecolumn
\section{Experimental details}
\label{app:exp-details}
Throughout our paper, we train various diffusion models on CIFAR-10 and MS~COCO.

\paragraph{DDPM training on CIFAR-10.} We train 100
DDPMs~\citep{ho2020denoising} on the  CIFAR-10 dataset for 200 epochs using a
cosine annealing learning rate schedule that starts at 1e-4. We used the DDPM
architecture that match the original implementation~\citep{ho2020denoising},
which can be found here \url{https://huggingface.co/google/ddpm-cifar10-32}. At
inference time we sample using a DDPM scheduler with 50 inference steps.

\paragraph{LDM training on MS~COCO.} We train 20 text-conditional latent
diffusion models (LDMs)~\citep{rombach2022high} on the MS~COCO dataset for 200
epochs using a cosine annealing learning rate schedule that ståarts at 2e-4. We
use the exact CLIP and VAE used in Stable Diffusion v2, but use a custom
(smaller) UNet, which we describe in our code. These models can be found
here~\url{https://huggingface.co/stabilityai/stable-diffusion-2-1}. At inference
time, we sample using a DDPM scheduler with 1000 inference steps.

\paragraph{LDS.} We sample 100 random $50\%$ subsets of CIFAR-10 and MS~COCO,
and train 5 models per mask. Given a set of attribution scores, we then compute
the Spearman rank correlation \citep{spearman1904proof} between the predicted
model outputs $g_\tau(\cdot)$ (see Eq.~(\ref{eq:data_attr_agg})) on each subset
according to the attributions and the (averaged) actual model outputs.  Because
our model output and attributions are specific to a step, we compute LDS
separately for each step. To evaluate the counterfactual significance of our
attributions over the course of the diffusion trajectory, we measure LDS scores
at each $100$ steps over the $1000$ step sampling process.

\paragraph{Retraining without the most influential images.}
For our counterfactual evaluation in \cref{sec:evaluation}, we compute
attribution scores on 50 samples from our CIFAR-10 and MS~COCO models at step $t
= 400$.  Given the attribution scores for each sample, we then retrain the model
after removing the corresponding top $k$ influencers for $k=200,500,1000$. We
compute FID based on $5000$ images from each distribution, and repeat this
process for each sample at each value of $k$.

\paragraph{\trak hyperparameters}
In the random projection step of \trak, we use a projection dimension of
$d=4096$ for CIFAR-10 and $d=16384$ for MS~COCO. As in \citet{park2023trak}, we
use multiple model checkpoints in order to compute the attribution scores. For
CIFAR-10, we use 100 checkpoints, and for MS~COCO, we use 20 checkpoints. In our code
repository
(\href{https://github.com/MadryLab/journey-TRAK}{github.com/MadryLab/journey-TRAK}),
we release the pre-computed \trak features for all of our models, allowing for a
quick computation of \trak scores on new synthesized images. In
\cref{eqn:scoring}, we use $k=20$ for both CIFAR-10 and MS~COCO.

\clearpage
\section{Additional Analysis and Results}
\subsection{Diffusion models are consistent across seeds}
\label{app:seed-consistency}
A priori, two independent models trained on the same dataset do not share the same latent space. That is, a given noise sequence $\beps_T,...,\beps_0$  could be denoised to two unrelated images for two different models.
However, we find empirically that latent spaces from two diffusion models are highly aligned; we call this property {\em seed consistency}. In fact, we find that images generated by many independently trained DDPMs on CIFAR-10 from the same random seed and nearly indistinguishable (see Figure \ref{appfig:seed_consist_cifar}, right). To evaluate seed consistency quantitatively, we measure the $\ell_2$ distance between images generated by two models when using identical or distinct noise sequences, and find that matching the noise sequences leads to a far smaller $\ell_2$ distances (see Figure \ref{appfig:seed_consist_cifar}, left).

We additionally evaluate seed consistency on multiple checkpoints of Stable
Diffusion (we use checkpoints provided at
\url{https://huggingface.co/CompVis/stable-diffusion}
and
\url{https://huggingface.co/runwayml/stable-diffusion-v1-5})
and find that images generated across these models with a fixed seed share
significantly more visual similarity that those generated from independent
random seeds (see Figure \ref{appfig:seed_consist_laion}.)

We take advantage of this property when evaluating the counterfactual impact of
removing the training examples relevant to a given generated image (see
\cref{sub:forgetting}).  Specifically, we now expect that retraining a model on
the full training set and then sampling from the same seed should produce a
highly similar image to the generated image of interest. Thus, we can evaluate
the counterfactual significance of removing the training examples with the top
attribution scores for a given generated image by retraining and measuring the
distance (in pixel space) of an image synthesized with the same seed to the
original generated image.

\begin{figure*}[!htbp]
    \centering
    \includegraphics[width=.49\linewidth]{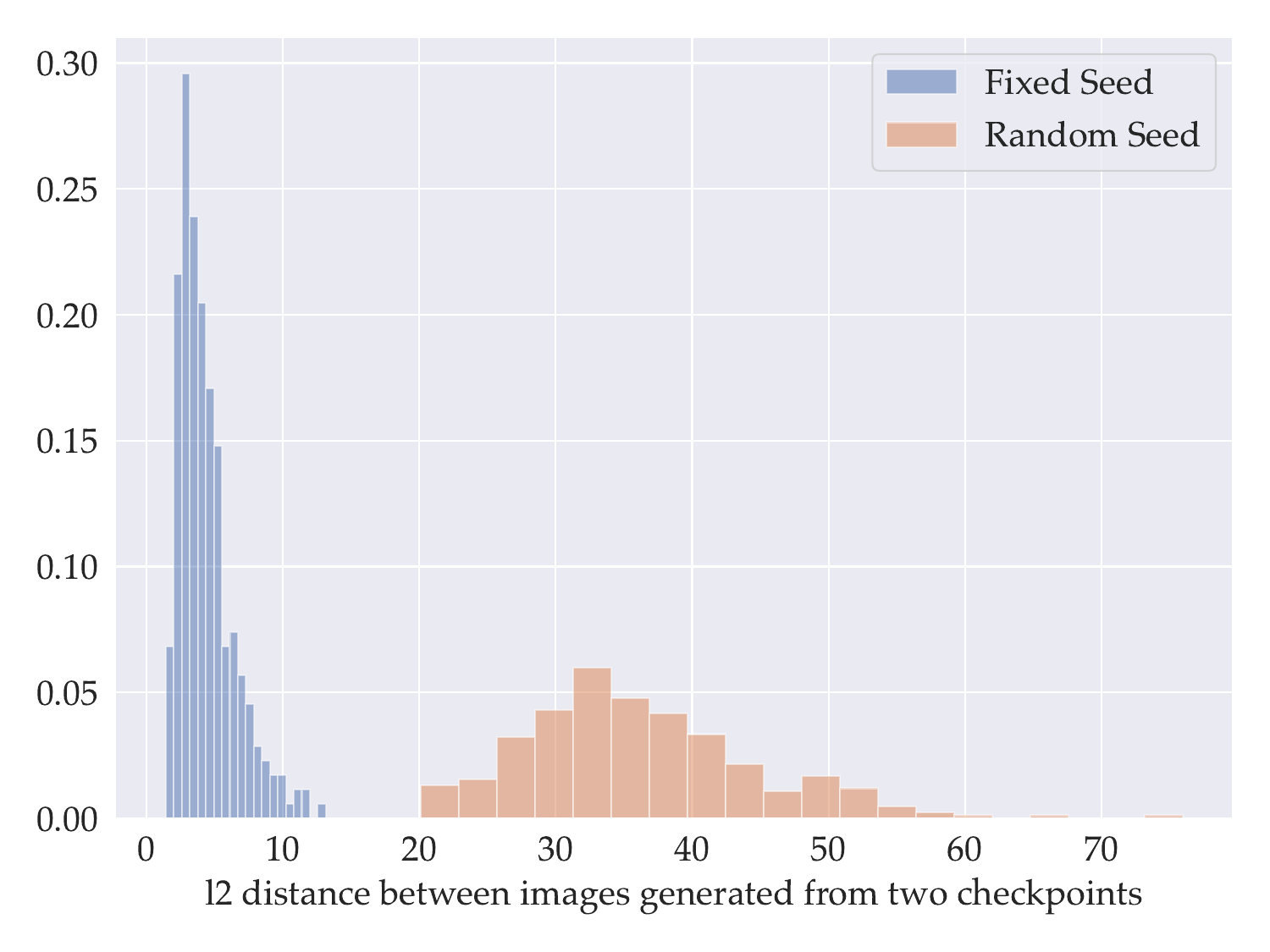}
    \hspace{0.8em}
    \includegraphics[width=.45\linewidth]{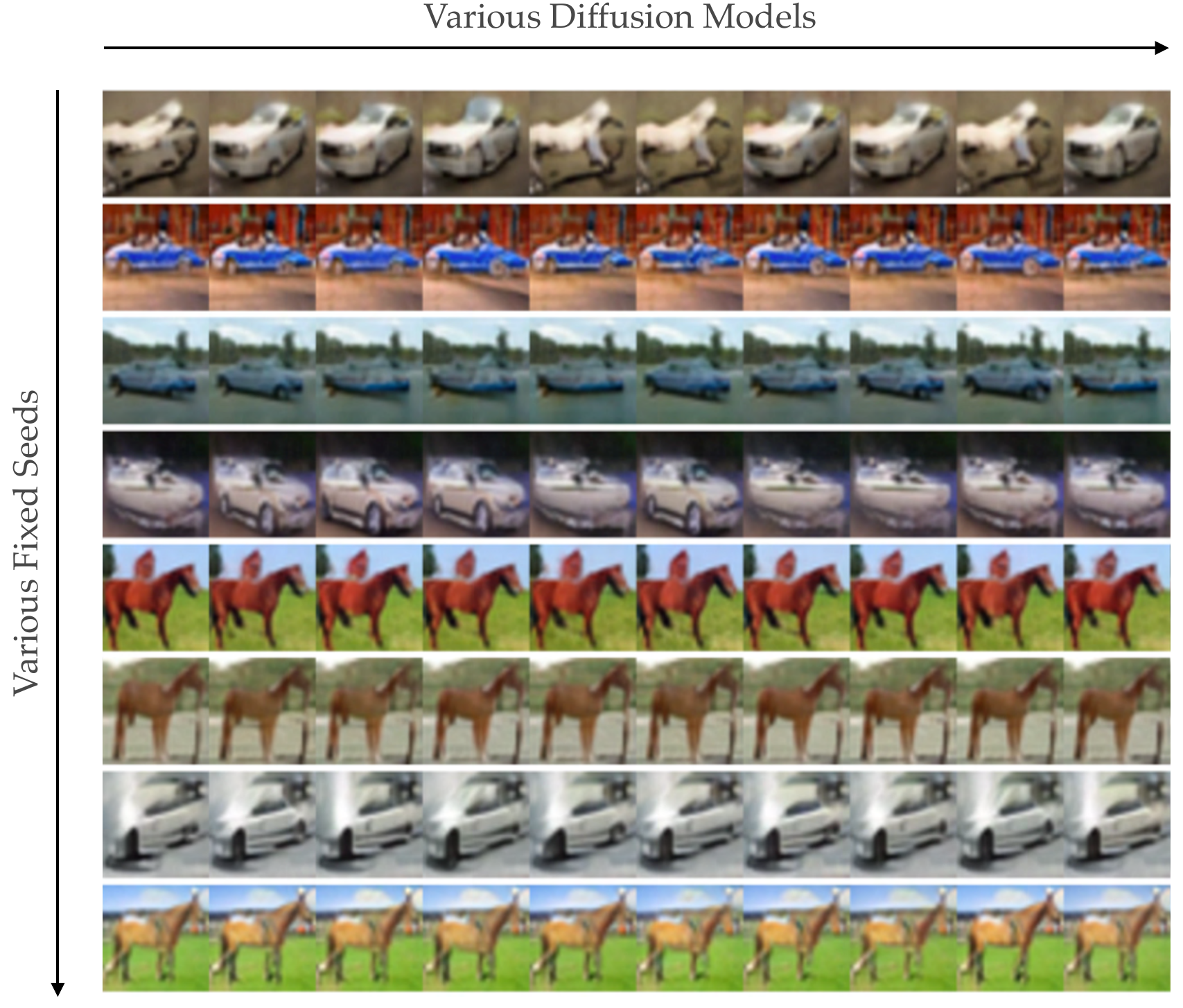}
    \caption{\textbf{Seed consistency of CIFAR-10 DDPMs}. We find that across DDPMs trained independently on CIFAR-10, when using a fixed random seed during sampling, the resulting synthesized images are very similar, and often visually indistinguishable \textbf{(Right)}. Quantitatively, we find that the $\ell_2$ distance between images generated from two different models is significantly smaller when we fix the noise sequence \textbf{(Left)}.}
    \label{appfig:seed_consist_cifar}
\end{figure*}

\begin{figure*}[!htbp]
    \centering
    \includegraphics[width=\linewidth]{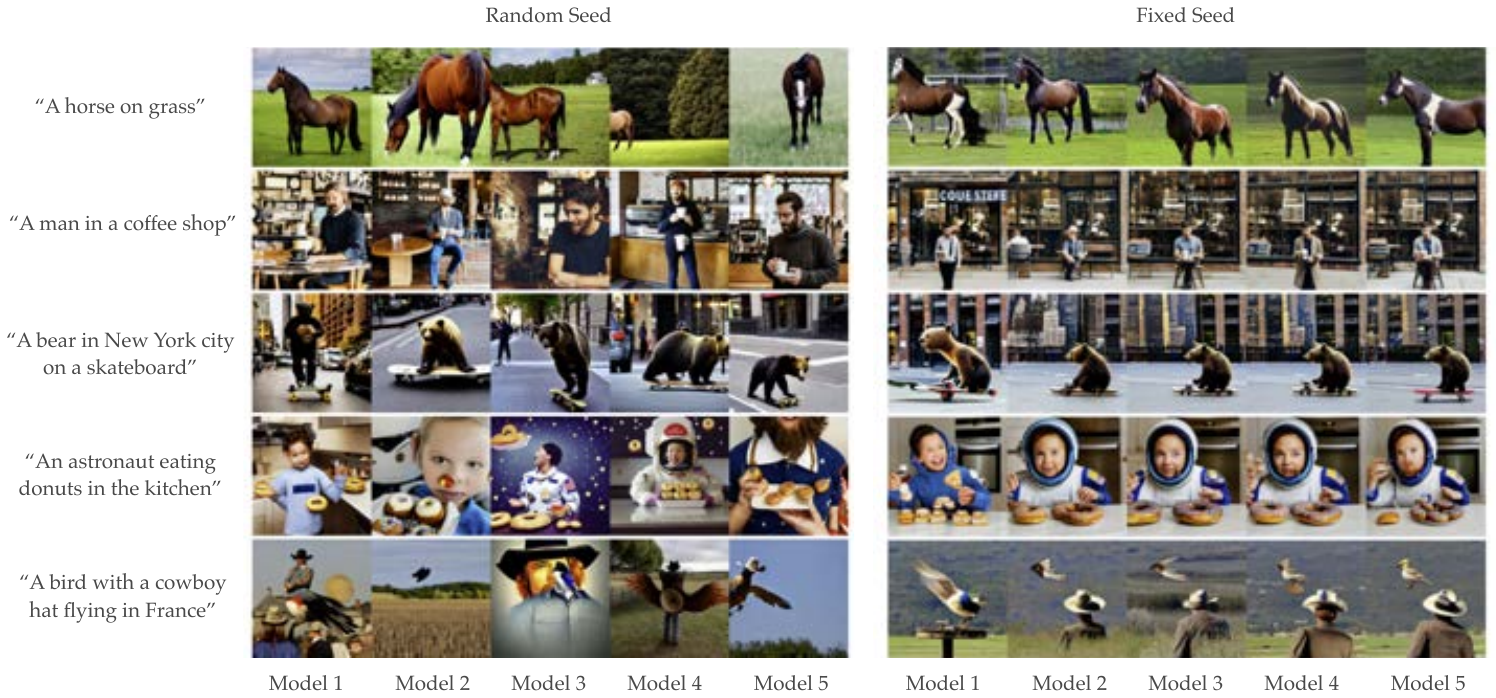}
    \caption{\textbf{Seed consistency holds for Stable Diffusion models.} We find that seed consistency holds even for large, text conditioned model, specifically for Stable Diffusion models that are trained on LAION-5B. We compare multiple checkpoints of Stable Diffusion provided by Stability AI, and find that fixing the noise sequence during sampling surfaces very similar images (in comparison to using independent noising sequences).}
    \label{appfig:seed_consist_laion}
\end{figure*}

\clearpage
\subsection{Attribution scores can drastically change over the course of the
diffusion process}
\label{app:why-per-step}
As additional motivation for performing attribution at individual steps rather
than the entire diffusion trajectory, we highlight the following phenomena:
\textit{the same training image can be both positively influential and negatively
influential for a generated sample at different steps}. For example, consider an
image of a red car on a grey background generated by our DDPM trained on
CIFAR-10 (See Figure \ref{appfig:why-not-avg}, top). We find that a specific
training example of a red car on grass is the single most positively influential
image according to \trak at the early stages of the generative process (as it is
forming the shape of the car), but is later the single most negatively
influential image (possibly due to the difference in background, which could
steer the model in a different direction). If we were to create an aggregate
attribution score for the entire diffusion trajectory, it is unclear what the
attribution score would signify for this training example.

To evaluate this phenomena quantitatively, we measure the percentage of generated images for which, for a given $K$, there exists a training example that is one of the top $K$ highest scoring images at some step and one of the top $K$ lowest scoring images at another step (according to \trak). In Figure \ref{appfig:neg-and-pos}, we show how this percentage varies with $K$. As a baseline, we also include the probability of such a training example existing given completely random attribution scores. We find that our observed probabilities match those expected with random scores, signifying that an image being highly positively influential at a given step \textit{does not} decrease the probability that it is highly negatively influential at a different step.

To more broadly analyze the relationship between attributions at different steps, we additionally measure the Spearman's rank correlation
\citep{spearman1904proof} between attribution scores for the same generated sample at different steps (see Figure \ref{appfig:steps_corr}). We find that for steps that are sufficiently far from each other (around 500 steps), the attribution scores are nearly uncorrelated.

\begin{figure*}[!htbp]
    \centering
    \includegraphics[width=\linewidth]{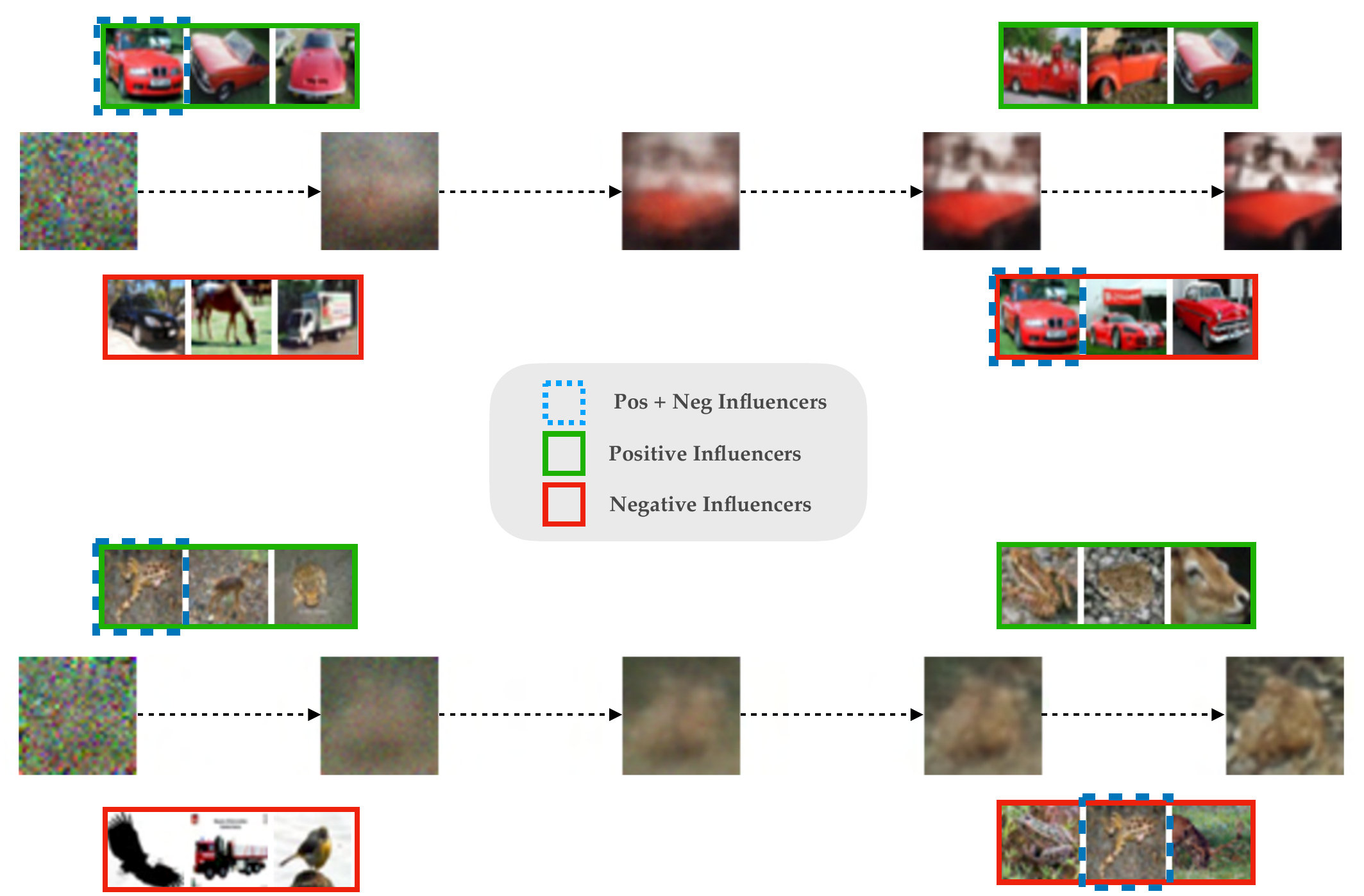}
    \caption{\textbf{Overlap between positive and negative influencers.} Here, we visualize the generative process for two images generated by a DDPM on CIFAR for which there exists a training image that is both positively and negatively influential at different steps. If we consider an aggregate attribution score across all time-steps of the diffusion trajectory, we might lose the significance of such training examples which alternate between being positively and negatively influential during the sampling process.}
    \label{appfig:why-not-avg}
\end{figure*}

\begin{figure*}
    \centering
    \includegraphics[width=0.6\linewidth]{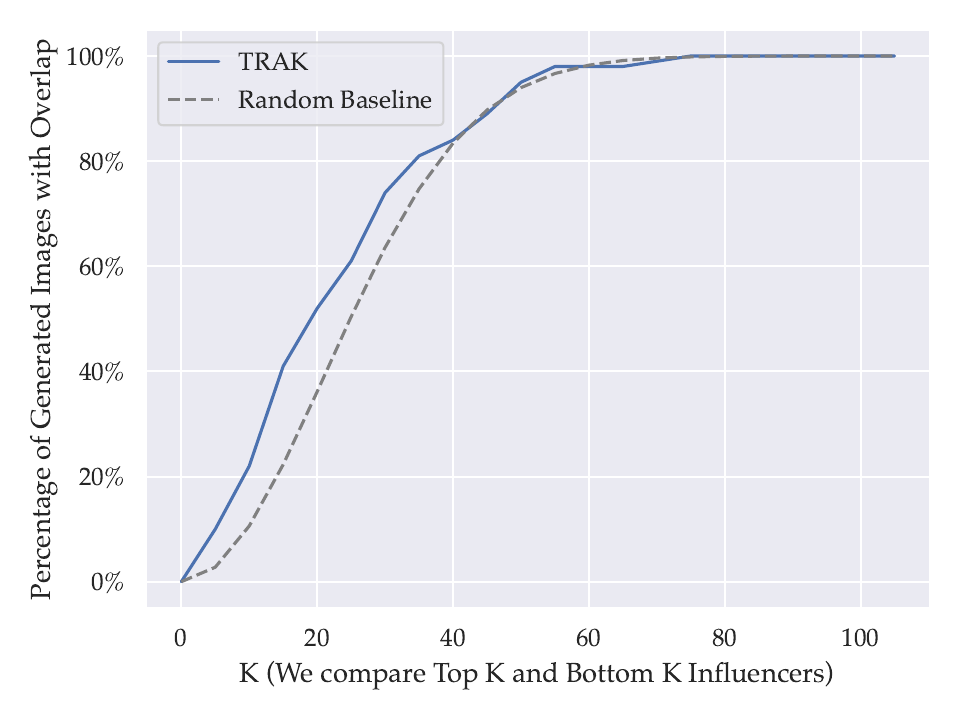}
    \caption{\textbf{The relationship between positive and negative influencers.} Here, we plot the probability that within the attribution scores for a given generated image, there exists a training example that is one of the $K$ most positive influencers at some step and one of the bottom $K$ most negative influencers at another step. We compute this probability empirically with the attribution scores from \trak and find that it closely aligns with the hypothetical baseline of completely random attribution scores. This signifies that being a top positive influencer at some step does not decrease the likelihood of being a top negative influencer at a different step.}
    \label{appfig:neg-and-pos}
\end{figure*}

\begin{figure*}
    \centering
    \includegraphics[width=0.6\linewidth]{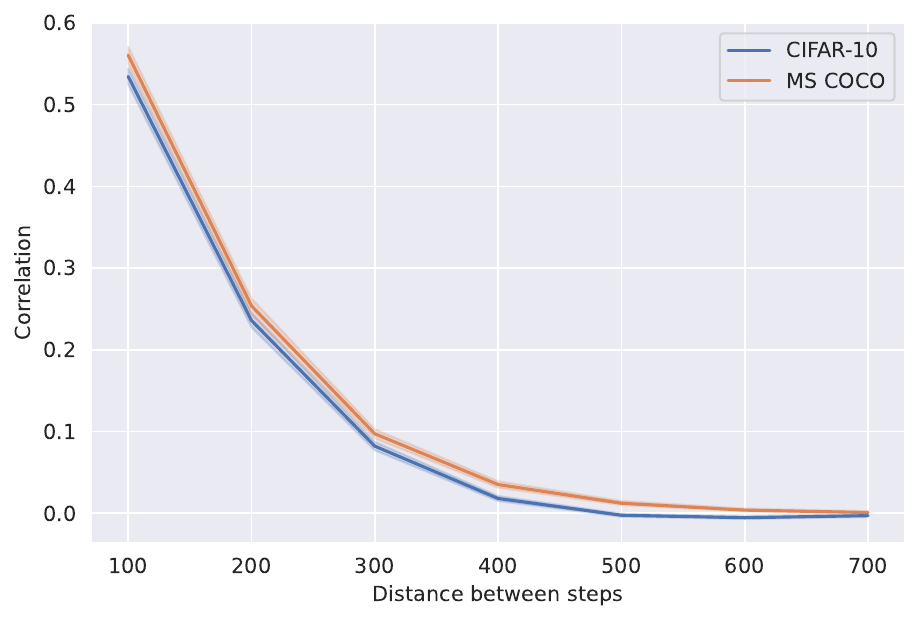}
    \caption{\textbf{Correlation between attribution scores over steps.} Here, we plot the Spearman's rank correlation
    \citep{spearman1904proof} between the attribution scores for a given image generated by either our CIFAR-10 or MS~COCO models at different steps, as a function of the distance between steps (results are averaged over 100 generated samples). As expected, steps that are closer in proximity have more closely aligned attribution scores. Interestingly, when we compute attributions at steps of distance 500 or more apart, the resulting scores are nearly uncorrelated.}
    \label{appfig:steps_corr}
\end{figure*}

\clearpage
\subsection{Feature analysis for Stable Diffusion}
\label{app:features_stable}
We analyze how the likelihood of different features in the final image varies over steps for images generated by a Stable Diffusion model,\footnote{We use the \texttt{stabilityai/stable-diffusion-2} pre-trained checkpoint.} similarly as we did for CIFAR-10 in \Cref{fig:phase_transition}.
In \Cref{fig:SD}, we analyze an image generated using the prompt, \emph{``A woman sitting on a unique chair beside a vase.''}
To measure the relative likelihood between two features (e.g., ``white blouse'' vs. ``dark blouse''), we use a pre-trained CLIP model and measure whether the CLIP embedding of the generated image is closer to the text embedding of the first feature or the second feature. We sample 60 images at each step and report the average likelihood.
We use 300 denoising steps to speed up the generation.

\begin{figure}[!h]
    \centering
    \includegraphics[width=1\textwidth]{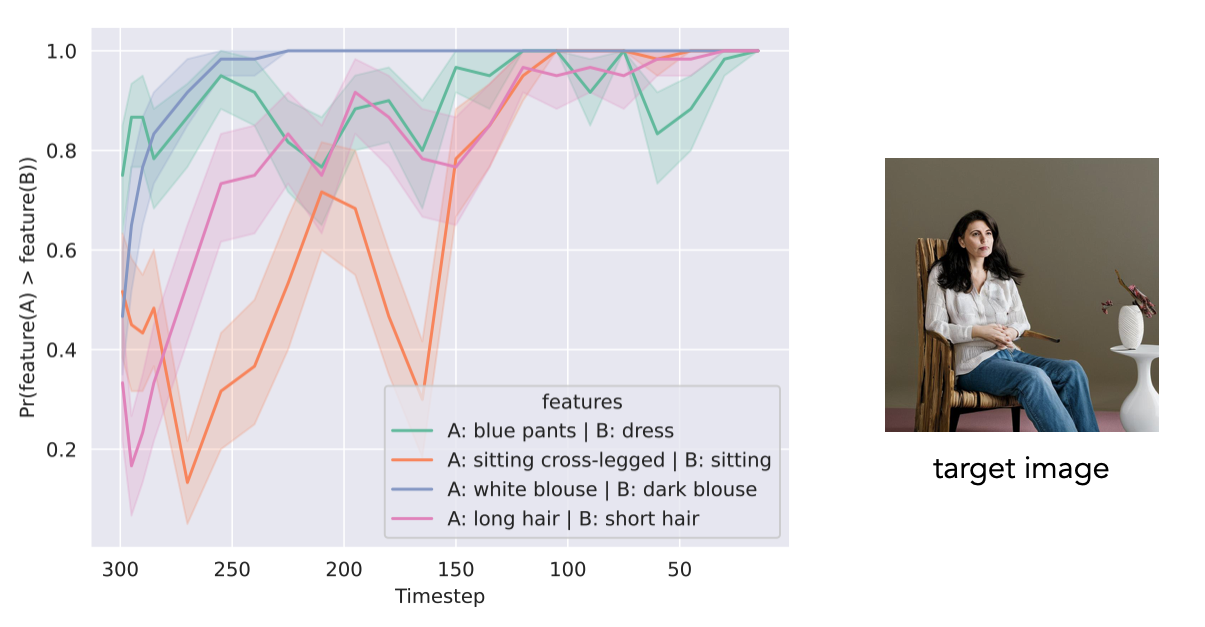}
    \vskip -.1cm
    \caption{
        \textbf{Features appear at specific steps for Stable Diffusion.} \textbf{(Left)} For each pair of features, we plot the evolution in the relative likelihood of the two features (according to CLIP text-image similarity) in the conditional distribution $p_{\theta}(\cdot|\bx_t)$. Features differ in when they appear, but usually rapidly appear within a short  interval of steps.
        (\textbf{Right}) The generated image $\bx_0$, sampled using $T=300$ denoising steps.
    }
    \label{fig:SD}
\end{figure}

\clearpage
\subsection{Omitted plots}
\label{app:omitted}
In this section, we present additional visualizations extending upon the figures in the main text. In \cref{appfig:traj-cifar}
and \cref{appfig:traj-coco}, we visualize the most influential training examples identified by our method for a sample generated with a DDPM trained on CIFAR-10 and a LDM trained on MS~COCO, respectively. In \cref{appfig:more-like-mainfig}, we more concisely display attributions for additional samples generated by a CIFAR-10 DDPM. Finally, in \cref{appfig:phase-classifier} we display additional examples of the appearance of features over steps, and confirm that our findings in the main text hold across when different classification models are used for identifying a given feature.

\begin{figure}[h]
    \centering
    \includegraphics[width=\linewidth]{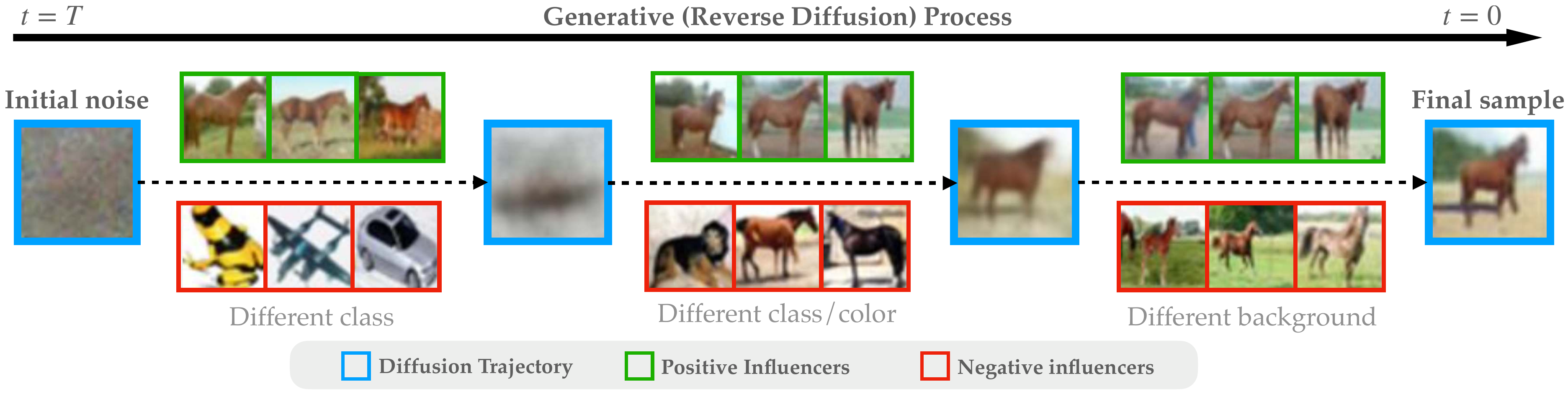}
    \caption{An example of step-dependent attribution scores for a sample generated by  a DDPM trained on CIFAR-10. At each step $t$, our method pinpoints
    the training examples with the highest influence (positive in {green},
    negative in red) on the generative process at this step. In particular,
    positive influencers guide the trajectory towards the final sample, while negative
    influencers guide the trajectory away from it.}
    \label{appfig:traj-cifar}
\end{figure}

\begin{figure}[h]
    \centering
    \includegraphics[width=\linewidth]{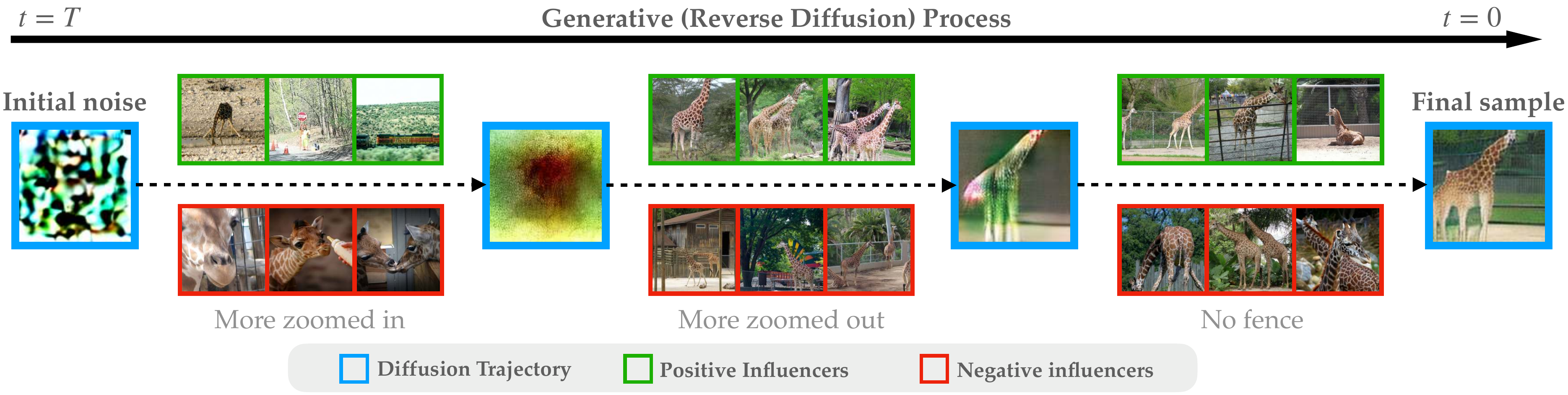}
    \caption{An additional example of step-dependent attribution scores for a sample generated by  a LDM trained on MS~COCO. At each step $t$, our method pinpoints
    the training examples with the highest influence (positive in {green},
    negative in red) on the generative process at this step. In particular,
    positive influencers guide the trajectory towards the final sample, while negative
    influencers guide the trajectory away from it.}
    \label{appfig:traj-coco}
\end{figure}

\begin{figure}[h]
    \centering
    \includegraphics[width=\linewidth]{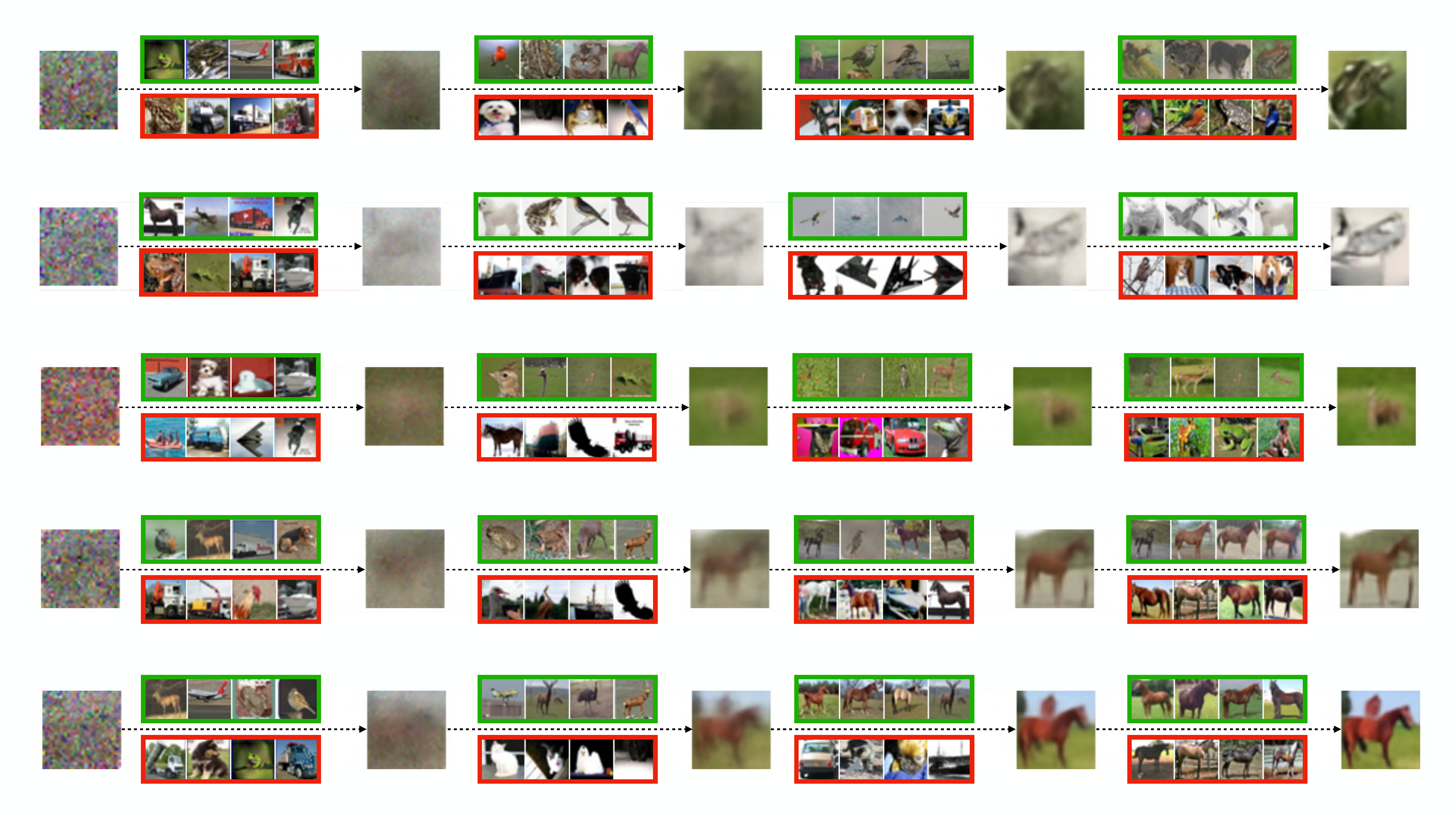}
    \caption{Additional examples our attributions identified by our method. Here, we visualize the diffusion trajectory for generated images along with the most positively (green) and negatively (red) influential images at individual steps throughout the diffusion trajectory.}
    \label{appfig:more-like-mainfig}
\end{figure}

\begin{figure}[h]
    \centering
    \includegraphics[width=\linewidth]{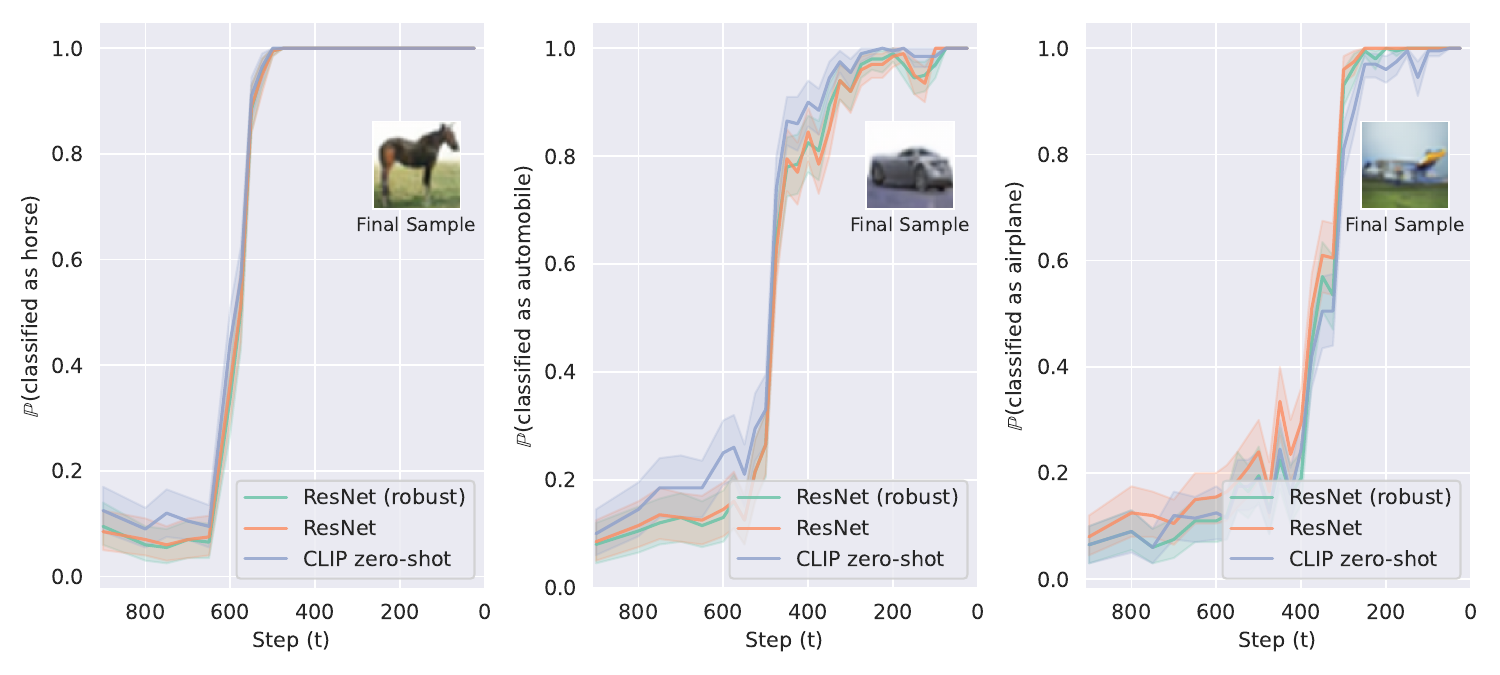}
    \caption{Additional examples of the appearance of features over steps, similar to the analysis in \cref{fig:phase_transition}. In each plot, we show the likelihood that a sample generated from the distribution $p_\theta(\cdot|\bx_t)$ contains a the feature of interest (in this case, the CIFAR-10 class of the final image) according to three different classifiers: a ResNet trained on the CIFAR-10 dataset with either standard or robust training, and zero-shot CLIP-H/14 model~\citep{radford2021learning}. Note that in each example, the likelihood that the final image contains the given feature increases rapidly in a short interval of steps, and that this phenomena is consistent across different classifiers.}
    \label{appfig:phase-classifier}
\end{figure}

\end{document}